\documentclass[journal,draftcls,onecolumn]{IEEEtran}
\usepackage{amsmath}
\usepackage{amssymb}
\usepackage{graphicx}
\usepackage{color}
\usepackage{cite}
\usepackage{bm}
\usepackage{epsfig,psfrag}
\usepackage{tabularx}
\usepackage{acronym}
\usepackage[yyyymmdd,hhmmss]{datetime}
\usepackage{bbm}
\usepackage{mathrsfs} 
\usepackage{bardiag}
\usepackage{tabularx}
\usepackage{multirow}
\usepackage{colortbl,pgfplotstable}
\usepackage{algorithm}
\usepackage{algpseudocode}
\usepackage{mathtools}
\usepackage{subfig}
\usepackage{setspace}
\usepackage{cite}
\usepackage{LatexInclusion/notation}
\usepackage[level=3]{LatexInclusion/messages}  % set level to 0, to hide all notes
\newcommand{\bd}{\begin{description}}
\newcommand{\ed}{\end{description}}
\newcommand{\be}{\begin{enumerate}}
\newcommand{\ee}{\end{enumerate}}
\newcommand{\bi}{\begin{itemize}}
\newcommand{\ei}{\end{itemize}}
\newcommand{\bl}{\begin{list}}
\newcommand{\el}{\end{list}}
\newcommand{\bt}{\begin{tabbing}}
\newcommand{\et}{\end{tabbing}}

\definecolor{BLUE}{rgb}{0,0,1}

\allowdisplaybreaks
\definecolor{BLUE}{rgb}{0,0,1}

\providecommand{\bl}{\textcolor{blue}}

\providecommand{\blk}{\textcolor{black}}

\providecommand{\ist}{\hspace*{.3mm}}

\providecommand{\rmv}{\hspace*{-.3mm}}

\providecommand{\rrmv}{\hspace*{-1mm}}
\providecommand{\nn}{\nonumber}

\providecommand{\T}{\mathrm{T}}

\acrodef{pdf}[PDF]{Probability Density Function}
\acrodef{mmse}[MMSE]{Minimum Mean-Square Error}
\acrodef{rmse}[RMSE]{Root Mean-Square Error}
\acrodef{mse}[MSE]{Mean-Squared Error}
\acrodef{pmf}[PMF]{Probability Mass Function}
\acrodef{sss}[SSS]{Side-scan Sonar}
\acrodef{auv}[AUV]{Autonomous Underwater Vehicle}
\acrodef{sauv}[sAUV]{Small Autonomous Underwater Vehicle}
\acrodef{asv}[ASV]{Autonomous Surface Vehicle}
\acrodef{uuv}[UUV]{Unmanned Underwater Vehicle}
\acrodef{pf}[PF]{Particle Filter}
\acrodef{ukf}[UKF]{Unscented Kalman Filter}
\acrodef{ekf}[EKF]{Extended Kalman Filter}
\acrodef{ins}[INS]{Inertial Navigation System}
\acrodef{dvl}[DVL]{Doppler Velocity Log}
\acrodef{cml}[CML]{Concurrent Mapping and Localization}
\acrodef{adcp}[ADCP]{Acoustic Doppler Current Profiler}
\acrodef{lbl}[LBL]{Long Baseline}
\acrodef{usbl}[USBL]{Ultra Short Baseline}
\acrodef{deif}[DEIF]{Distributed Extended Information Filter}
\acrodef{dr}[DR]{Dead Reckoning}
\acrodef{ins}[INS]{Inertial Navigation System}
\acrodef{slam}[SLAM]{Simultaneous Localization and Mapping}
\acrodef{cf}[CF]{Cumulative Frequency}

\acrodef{pda}[PDA]{Probabilistic Data Association}
\acrodef{phd}[PHD]{Probability Hypothesis Density}
\acrodef{iid}[iid]{independent and identically distributed}
\acrodef{imu}[IMU]{Inertial Measurement Unit}
\acrodef{mot}[MOT]{Multi-Object Tracking}
\acrodef{mou}[MOU]{Measurement-Origin Uncertainty}
\acrodef{isam}[iSAM]{Incremental Smoothing and Mapping}
\acrodef{jcbb}[JCBB]{Joint Compatibility Branch and Bound}

\definecolor{BLUE}{rgb}{0,0,1}

\definecolor{myred}{rgb}{1,0.27,0}
\definecolor{mygreen}{rgb}{0.1, 0.55, 0.1}
\definecolor{myblue}{rgb}{0, 0, 1}

\newdateformat{monthyeardate}{\monthname[\THEMONTH] \THEDAY, \THEYEAR} % specifies the format of the date

\newcommand{\paperTitle}{A \blk{Landmark-Aided} Navigation Approach\\ Using Side-Scan Sonar}

\pgfplotsset{compat=1.14}

\begin{document}
\title{\paperTitle \vspace*{2mm}}

\author{Ellen Davenport, Khoa Nguyen, Junsu Jang, Clair Ma, Sean Fish, Luc Lenain, and Florian Meyer \vspace{0mm}

\thanks{ The material presented in this work was supported by the Office of Naval Research under Grants No. N00014-23-1-2284, N00014-23-1-2243, and N00014-24-1-2796, as well as by the National Science Foundation under CAREER Award No. 2146261. Parts of this work have been presented at the FUSION-23, Charleston, SC, USA.
}

\thanks{Ellen Davenport, Khoa Nguyen, Junsu Jang, Clair Ma, Luc Lenain, and Florian Meyer are with the Scripps Institution of Oceanography and the Department of Electrical and Computer Engineering, University of California San Diego, La Jolla, CA 92093, USA (e-mail: \{davenport, ktnguyen, jujang, c2ma, llenain, flmeyer\}@ucsd.edu).}

\thanks{Sean Fish is with the Daniel Guggenheim School of Aerospace Engineering, Georgia Institute of Technology, Atlanta, GA 30332, USA (e-mail: seanfish@gatech.edu). This work was performed as a visiting student at the University of California San Diego, La Jolla, CA 92093.}

\vspace{-4mm}
}

\maketitle

% \doublespacing
% \linenumbers

\begin{abstract}
Cost-effective localization methods for \ac{auv} navigation are key for ocean monitoring and data collection at high resolution in time and space. Algorithmic solutions suitable for real-time processing  that handle nonlinear measurement models and different forms of measurement uncertainty will accelerate the development of field-ready technology. This paper details a Bayesian estimation method for \blk{landmark-aided} navigation using a \ac{sss} sensor. The method bounds navigation filter error in the GPS-denied undersea environment and captures the highly nonlinear nature of slant range measurements while remaining computationally tractable. Combining a novel measurement model with the chosen statistical framework facilitates the efficient use of \ac{sss} data and, in the future, could be used in real time. The proposed filter has two primary steps: a prediction step using an unscented transform and an update step utilizing particles. The update step performs probabilistic association of sonar detections with known landmarks. \blk{We evaluate algorithm performance and tractability using synthetic data and real data collected field experiments. Field experiments were performed using two different marine robotic platforms with two different \ac{sss} and at two different sites. Finally, we discuss the computational requirements of the proposed method and how it extends to real-time applications.}
\end{abstract}

\begin{IEEEkeywords}
Side-scan sonar, Bayesian estimation, autonomous vehicles, and probabilistic data association.
\vspace{0mm}
\end{IEEEkeywords}

\section{Introduction}
\label{sec:introduction}

Data collection and monitoring are notoriously difficult at sea due to the inherently high cost, large geographic scale, and potential risk to humans and equipment. As the demand for ocean data has increased, \acp{sauv} have come to the forefront in facilitating cheaper, safer, and more complex data collection operations. A key challenge in autonomous underwater navigation, however, is the availability of absolute positioning information. In the absence of GPS measurements, \acp{sauv} typically rely on dedicated sensors for \ac{dr} such as an \ac{imu} or a \ac{dvl}. However, the estimation error associated with these methods grows unbounded \cite{LeoBah:B16, PetBroPenSus:J18, RuiPetLan:C03, MicPra:C14}. Larger vehicles, such as submarines, address this issue with \acp{ins} that are cost and size-prohibitive for smaller platforms.

It is possible to constrain the vehicle location by surfacing to receive GPS signal, but this can reduce time underwater and restrict operational depth. Alternatively, one can use acoustic triangulation techniques such as \ac{lbl} and \ac{usbl} positioning systems \cite{RigPizWil:C06,PauSaeSetHow:J14, RogTho:J99,CheDonMilFar:J16}. The downside of these acoustic ranging methods is that they require the deployment of transponder infrastructure in the form of moorings, which are expensive and stationary, thus limiting the overall scope of the mission to a fixed region \cite{RuiRauPetLan:J04}. This highlights the need for \ac{sauv} navigation strategies that combine small, inexpensive onboard sensors with sophisticated signal processing methods to allow vehicles to remain underwater indefinitely without limiting their range. These improvements will expand the viability of small, cost-effective autonomous solutions in private and academic sectors.

\subsection{Navigation with \ac{sss}}

\blk{Landmark-aided} navigation is a promising research area that aims to bound the localization error using identifiable seabed features detected with onboard sensors \cite{RuiPetLan:C03, RuiRauPetLan:J04, FalKaeJohLeo:C11,PetBroPenSus:J18,MicPra:C14,PetReeBel:C02, StaBleUra:C08}. \blk{The landmark-aided approach is distinct from terrain-based or terrain-relative navigation, which utilizes estimates of vehicle height from the floor but does not use landmark information \cite{Anonsen:C06,Meduna:C08,MelMat:J17}. If landmark locations are unknown, they can potentially be estimated jointly with the location of the platform in a \ac{slam} approach \cite{RuiRauPetLan:J04,ReeRuiCapPet:J06,Sia:C16}.} Most landmark-aided navigation approaches, however, assume that the area of interest has been surveyed at least once, i.e., the \ac{sauv} has performed multiple passes \cite{RuiRauPetLan:J04}, the mission is supported by a lead vehicle \cite{FalKaeJohLeo:C11}, or previously generated seafloor maps are available. Our approach also follows this assumption.

\ac{sss} is a standard payload on \acp{sauv} and is used for applications such as hydrogeographic survey and monitoring of coastal environments \cite{RuiPetLan:C03, RuiRauPetLan:J04, FalKaeJohLeo:C11, PetBroPenSus:J18}. The sensor has a very small form factor and generates large, high-resolution images of the seafloor, making it well-suited for \blk{landmark-aided} navigation. \ac{sss} navigation has been attempted before \cite{Sia:C16,RuiRauPetLan:J04,FalKaeJohLeo:C11,WooFre:C10,StaBleUra:C08,Haraldstad:T23} but remains a challenging research problem because of the nonlinear relationship between sonar pings and detected landmarks in addition to the variability of sensor performance due to platform motion and the acoustic background noise \cite{PetReeBel:C02, ZhuIsaFuFer:C17}. 

\subsection{\ac{mou}}

Another key challenge in \blk{landmark-aided} navigation is \ac{mou}, i.e., the ambiguity regarding which known landmark generates each sonar detection. Performing \blk{landmark-aided} navigation in real-time requires a data association step that handles this uncertainty \cite{BarDauHua:J09, BarWilTia:B11}. Previous work in \cite{RuiRauPetLan:J04}, \cite{Sia:C16}, and \cite{FalKaeJohLeo:C11} assess the viability of \ac{sss} in a \ac{slam} framework. For simplicity, the first of these approaches does not include the data association step. The second method performs data association but acknowledges that the method fails in the case of a false detection. The work in \cite{FalKaeJohLeo:C11} extends previously introduced approaches by implementing \ac{slam} with a sequential smoothing filter called \ac{isam} but performs data association manually and offline. This existing work successfully shows that \ac{sss} landmark detections can be used for the positioning of an underwater vehicle. However, robust solutions for data association are still required for landmark-aided navigation in real-time.

There are many existing robotics and estimation applications that handle \ac{mou}. For example, methods that perform ``hard'' detection-to-landmark associations include global nearest neighbor, individual compatibility, and \ac{jcbb} \cite{MacSetPan:C15,BarWilTia:B11}. As mentioned above, \cite{Sia:C16} uses \ac{jcbb} to perform association of detected landmarks, but the method is sensitive to false detections. Algorithms such as \ac{pda} and multiple hypothesis tracking can increase robustness by performing either ``soft'' probabilistic associations \cite{BarDauHua:J09, BarWilTia:B11,ForBarSch:J83,SchBurFoxCre:C01} or ``hard'' associations over a sliding window of time \cite{Rei:J79,CorCar:J18}. Our localization algorithm includes a method for handling \ac{mou} based on \ac{pda}, which is robust to missed detections or incorrect associations and has recently been explored in underwater navigation with \ac{sss} \cite{Haraldstad:T23}.

\blk{\ac{pda} has been applied to address \ac{mou} in a handful of applications that use underwater sonar systems. For example, both \cite{Melo:C19} and \cite{Zacchini:J23} focus on the task of identifying and tracking objects. This method uses a forward looking sonar, and incorporates probabilistic data association with the goal of localizing the objects given their relationship to the vehicle. This application is useful for \ac{auv} tasks that require identification and labeling of specific objects on the sea floor, but does not localize the vehicle itself. In the future, this methodology could be combined with the navigation method proposed in this to build an evolving map of labeled objects.}

\subsection{Identification of Landmark Detections}
In much of the existing work discussed thus far, landmark detection is manual rather than automatic to emphasize the navigation problem rather than the image processing problem. \blk{After data collection, landmarks are manually identified offline as objects that appear on the seafloor and bounding boxes are placed around them. This does not exclude the possibility of clutter or missed detections as the \ac{sss} images can be noisy. Even when detecting landmarks through a manual process, the true landmark label is unknown. Manual detection is} also used in \cite{FalKaeJohLeo:C11} and \cite{RuiRauPetLan:J04}, both precursors to the work in this paper. Although automatic landmark detection is outside the scope of this paper, there is ongoing sonar image segmentation and object classification work that complements our research  \cite{PetBroPenSus:J18, KarFabBouAug:C06, RuiLanCha:J99, StaBleUra:C08, ShaSonGuoFenLiHeYan:C17, SonZhuGuaLiFenHeYan:C17, ReeRuiCapPet:J06}. \cite{ZhuIsaFuFer:C17} demonstrates a deep-learning solution for both target detection and classification. For our navigation algorithm to be implemented online, automatic landmark detection would be required, and classification or labeling could further improve the data association solution. 

\subsection{Contributions}

This paper addresses the fundamental problem of navigating \acp{sauv} in GPS-denied ocean environments while relying solely on cost-effective onboard sensors. We use detections of landmarks at known positions in a sequential Bayesian estimation \cite{AruMasGorCla:02,CagGerHod:J12,MeyBraWilHla:J17} method that is suitable for real-time applications and resource-constrained platforms. We address the computational complexity related to the processing of sonar measurements and the ambiguities inherent in landmark detections. In contrast to existing methods, we use each sonar ping as a separate measurement rather than concatenating pings into \ac{sss} images \cite{RuiLanCha:J99, RuiRauPetLan:J04, FalKaeJohLeo:C11, PetBroPenSus:J18, Haraldstad:T23}. In this way, our method builds on and branches from the one in \cite{Haraldstad:T23} by simplifying the measurement model and utilizing individual pings, both of which aid in increasing the efficiency of the filter. We incorporate ideas from \cite{MeyBraWilHla:J17,JawMihCanBul:C06} where \ac{pda} is embedded directly in a particle filter and from \cite{MeyBraWilHla:J17,MeyKroWilLauHlaBraWin:J18,JanMeySny:J23} where \ac{pda} is performed using belief propagation to improve scalability and reduce runtime.

\begin{figure*}[!htb]
    \centering
    \subfloat[]{ \includegraphics[scale=.06]{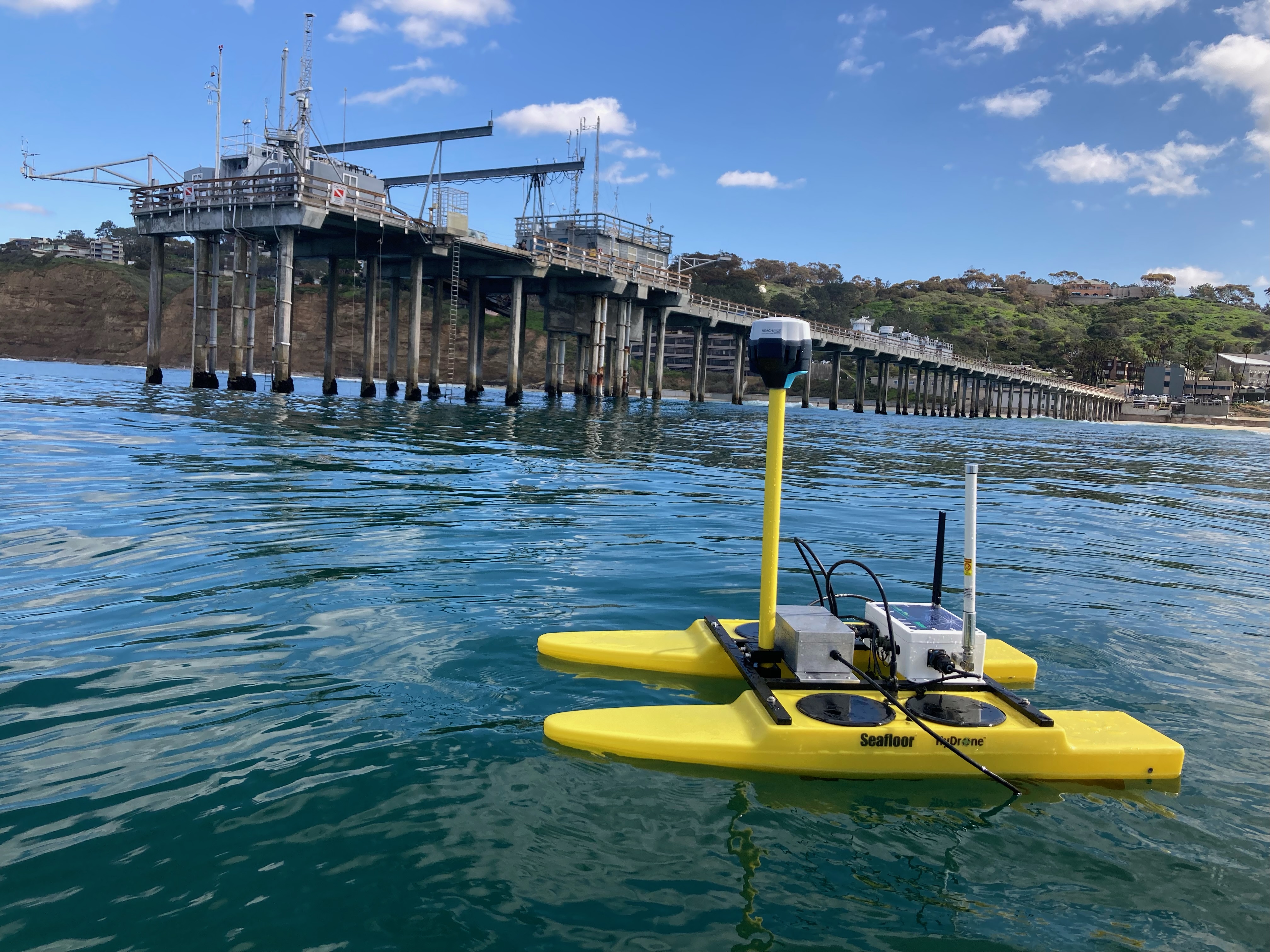} } \hspace{1mm}
    \subfloat[]{ \includegraphics[scale=.2915]{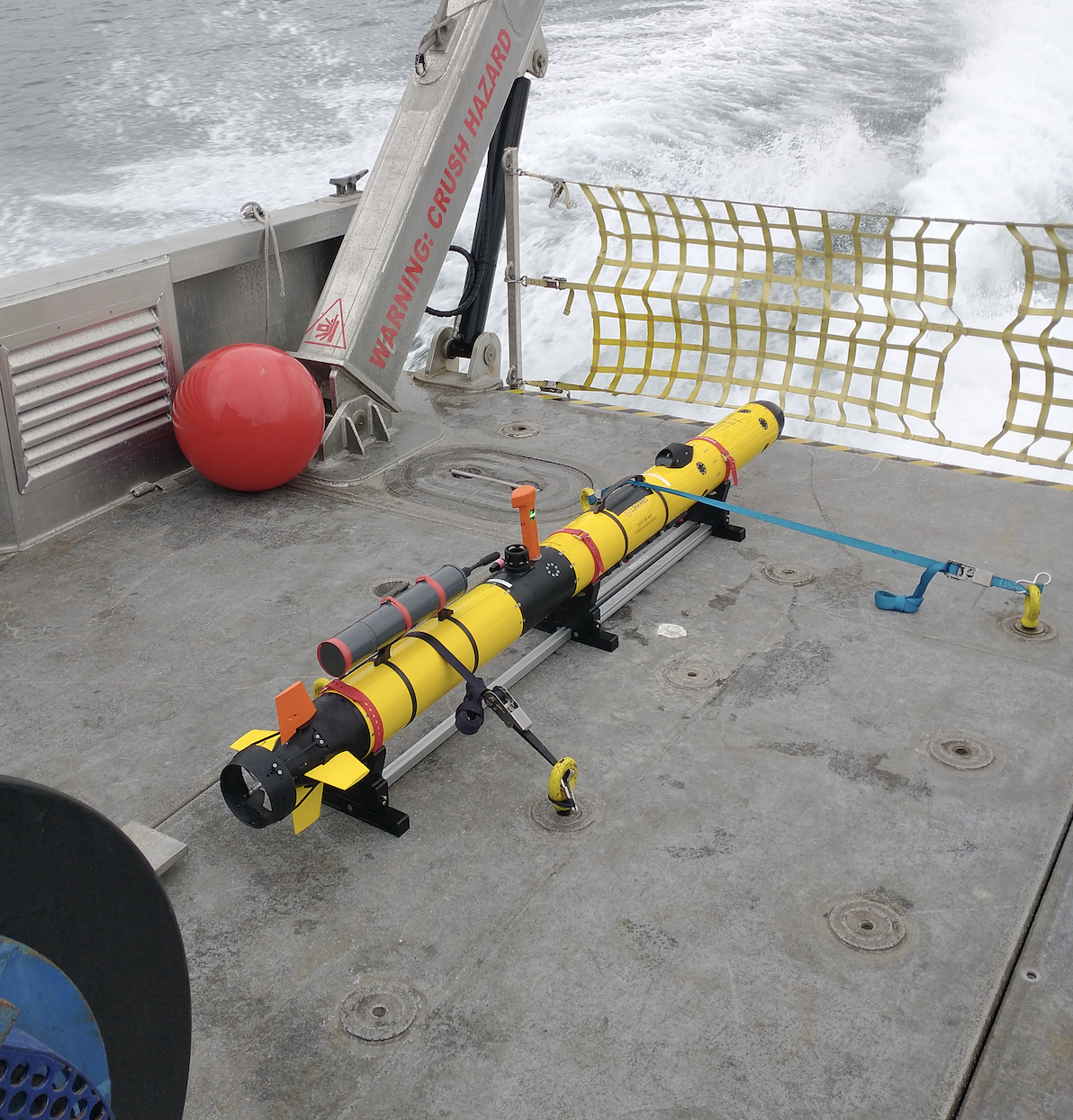} }
    \caption{\small Hydrone surface platform (a) and Iver3 underwater vehicle (b) used for \ac{sss} data collection and performance evaluation.}
    \label{fig:platforms}
    \vspace{-2mm}
\end{figure*}

The proposed method consists of a prediction step using an unscented transform \cite{JulUhl:04} and an update step with data association utilizing particles \cite{AruMasGorCla:02}. We evaluated the proposed method in simulation and in the field with data collected by a surface vehicle in Mission Bay, CA \blk{and by an underwater vehicle in La Jolla, CA}. The two platforms used for data collection are shown in Fig.~\ref{fig:platforms}. This paper discusses the quality and characteristics of the collected sonar data and algorithm performance in simulation and real-world scenarios. The key contributions of this paper can be summarized as\vspace{.8mm}:
\begin{itemize}
\item We establish a measurement model for \ac{sss} that relies on individual returns rather than entire \ac{sss} images.
\vspace{.8mm}
\item We develop a sequential Bayesian estimation method for \blk{landmark-aided} navigation using \ac{sss} landmarks.
\vspace{.8mm}
\item We include a probabilistic association of detections that is robust to false positives and missed landmarks\vspace{.8mm}.
\item We evaluate the resulting navigation filter based on simulated data and on data collected at sea. 
\vspace{.8mm}
\end{itemize}

This paper advances over the preliminary account of our method provided in the conference publication \cite{DavJanMey:C23} by (i) introducing a more advanced SSS measurement model that accounts for \ac{mou}; (ii) extending simulation results, and (iii) applying the resulting sequential Bayesian estimation method to field data. The remaining sections are organized as follows. Section \ref{sec:SystemModel} describes the state transition and measurement models. Section \ref{sec:measurementsAndMOU} discusses handling \ac{mou}. Section \ref{sec:Methods} presents the proposed navigation filter \blk{and discusses scalability}. Section \ref{sec:DataCollect} describes simulation, data collection, and preprocessing. Finally, Section \ref{sec:Results} presents numerical results from simulation and field experiments that demonstrate the feasibility and effectiveness of our algorithm. \blk{Section \ref{sec:Results} also includes a discussion of computational requirements}. Section \vspace{-.5mm}\ref{sec:Conclusion} concludes our work and discusses future research.

\section{Motion Model and General Measurement Model}
\label{sec:SystemModel}

In this section, we describe the vehicle motion model and measurement model for \ac{sss} pings.

\subsection{Motion Model}

At discrete time step $k$, the \ac{sauv} state is defined as $\V x_k \rmv=\rmv [ x_k  \ist\ist\ist y_k  \ist\ist\ist \theta_k \ist\ist\ist \gamma_k ]^{\T}$ where $[x_k \ist\ist\ist y_k]^{\T}$ is the 2-D position in a Cartesian coordinate system. We define $\theta_k$ as the heading in radians and $\gamma_k$ as the altitude above the seafloor. The control input vector is defined as $\V{u}_k \rmv=\rmv [u_{\mathrm{s},k} \ist\ist\ist u_{\mathrm{t},k} ]^{\T}\rmv\rmv$ where the first and second term denote speed and turn rate respectively. The nonlinear model $\V x_k \rmv=\rmv g(\V x_{k-1}, \V n_{k}, \V u_k)$ describes the transition of vehicle state from time $k \rmv-\rmv 1$ to time $k$. Additionally, the model incorporates a vector of driving noise terms $\V n_k \rmv=\rmv [ n_{\mathrm{s},k}  \ist\ist\ist n_{\mathrm{t},k} \ist\ist\ist n_{\theta,k}  \ist\ist\ist n_{\gamma,k} ]^{\T}\rmv\rmv$. Each element of driving noise is zero-mean, statistically independent, and Gaussian distributed with respective variances $\sigma^2_{\mathrm{s}}$, $\sigma^2_{\mathrm{t}}$, $\sigma^2_{\theta}$, and $\sigma^2_{\gamma}$. Driving noise terms are also assumed to be statistically independent across time. The functional form of the state transition model, $g(\V x_{k-1}, \V n_{k}, \V u_k)$,  is given\vspace{2mm} by
\begin{align}
\begin{bmatrix}
       x_k  \\[0.3em]
       y_k \\[0.3em]
      \theta_k  \\[0.3em]
      \gamma_k
\end{bmatrix} \rmv\rmv=\rmv\rmv \begin{bmatrix}
       x_{k-1} - \frac{v_{\mathrm{s},k}}{v_{\mathrm{t},k}}\sin(\theta_{k-1}) + \frac{v_{\mathrm{s},k}}{v_{\mathrm{t},k}}\sin(\theta_{k-1} + v_{\mathrm{t},k} \ist \Delta_k)\\[0.3em]
     y_{k-1} + \frac{v_{\mathrm{s},k}}{v_{\mathrm{t},k}}\cos(\theta_{k-1}) - \frac{v_{\mathrm{s},k}}{v_{\mathrm{t},k}}\cos(\theta_{k-1} + v_{\mathrm{t},k} \ist \Delta_k)\\[0.3em]
     \theta_{k-1} + v_{\mathrm{t},k} \ist \Delta_k + n_{\theta,k} \ist \Delta_k\\[0.3em]
      \gamma_{k-1} + n_{\gamma,k}
     \end{bmatrix}.
     \label{eq:stateTransitionModel}
     \vspace{1mm}
\end{align}
Here we introduce the notation for noisy speed and turn rate as $v_{\mathrm{s},k} = u_{\mathrm{s},k} + n_{\mathrm{s},k} $ and  $v_{\mathrm{t},k} = u_{\mathrm{t},k} + n_{\mathrm{t},k} $. $\Delta_k$ is the length of a discrete time step. The transition model for $[ x_k  \ist\ist\ist y_k  \ist\ist\ist \theta_k]^{\T}$ was developed in \cite{ThrFoxBur:B05}. The updated vehicle altitude is simply the altitude from the previous time step corrupted by additive noise.

\subsection{\ac{sss} Sonar Measurements}
\label{sec:SSSmodel}

\begin{figure}[h!]
\begin{center}
  \includegraphics[scale=.65]{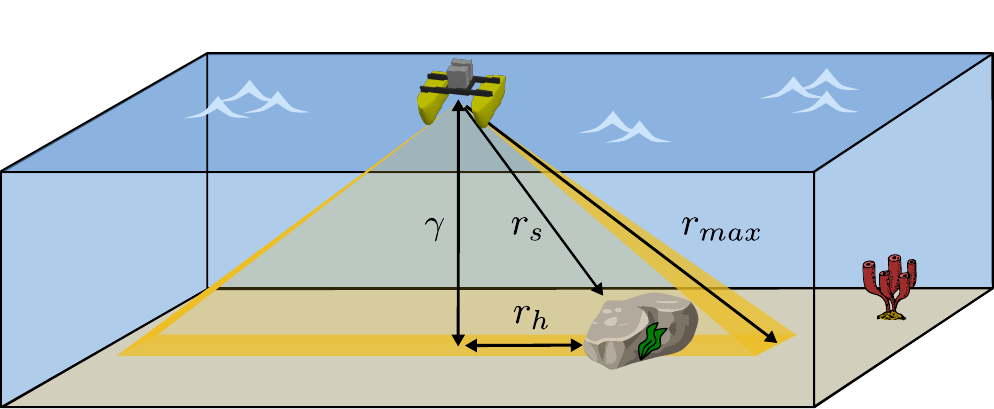}
  \vspace{-.7mm}
\end{center}
  \caption{A \ac{sss} sensor takes cross-track measurements. The maximum range of the SSS sensor, $r_{\mathrm{max}}$, shows the farthest point observed by the \ac{sss}. When a landmark is present, its respective slant range, $r_s$, can be extracted from the \ac{sss} image. This slant range corresponds to a horizontal distance, $r_h$. The vehicle altitude above the seafloor, $\gamma$, is the third side of this triangle and is needed to obtain $r_h$ from $r_s$.}
\label{fig:sonarGeo}
\end{figure}

\ac{sss} transducers send out acoustic pulses (``pings'') and generate an image from backscatter caused by features on the seafloor. Rough features generate stronger backscatter than smooth features, meaning certain materials have stronger acoustic signatures than others \cite{Hodges:B10}. Each ping results in an observation that corresponds to a new line of pixels in the \ac{sss} image. Observations are perpendicular to the direction of motion and thus are referred to as ``cross-track'' measurements. A \ac{sss} sensor generates an image by stitching observations from each time step $k$ in the direction of motion. 

In most cases, two \ac{sss} transducers are mounted in parallel so that observations can be simultaneously performed on the port and starboard side of the platform. Fig.~\ref{fig:sonarGeo} shows the geometry relating a \ac{sss} sensor to a surface vehicle, the water column, and the seafloor. Each pixel in an image will correspond to backscatter intensity within a particular region on the seafloor. Because the scenario is 3-dimensional, the distance to each pixel is called the ``slant range'' and represents distance through the water column (not along the seabed) \cite{CerMou:J93}. The size of the area corresponding to each pixel and the effective range of the \ac{sss} depends on the hardware configuration (e.g., transmit frequency), vehicle speed, and altitude above the seafloor. The random variability of acoustic propagation, including changes in the sea state or bottom topography \cite{Hodges:B10}, affects the quality of the image.

\begin{figure}[h!]
\begin{center}
  \includegraphics[scale=.5]{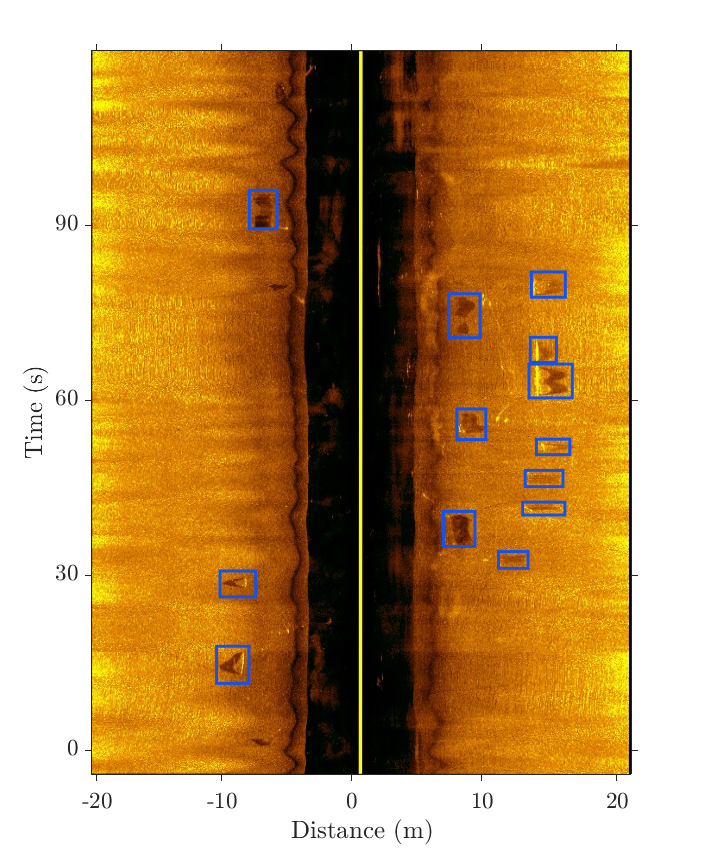}
    \vspace{-2mm}
\end{center}
  \caption{\ac{sss} image with landmarks annotated by boxes. The horizontal axis shows the slant range. Negative distance corresponds to the port side, and positive distances correspond to the starboard side. This SSS image shows approximately 2 minutes of data, and the maximum slant range at each side is 20m. The center frequency of the transmitted pulse is $1800$ kHz, and the ping rate is 30 Hz, meaning that each line of pixels is about 33 ms apart. The dark section at the center of the image is referred to as the nadir. This data was collected at the \textit{Scripps Institute of Oceanography} using the sensing platform shown in Fig.~4.}
  \label{fig:sonarImage}
  \vspace{-3mm}
\end{figure}

Fig.~\ref{fig:sonarImage} shows an example image with many identifiable landmarks. Each acoustic ping takes some time to travel through the water column, generating a region of dark pixels at the center called the \textit{nadir}. The width of the nadir is proportional to the time of the first acoustic return, often referred to as the first bottom return, and can be used to measure vehicle altitude above the seafloor \cite{RawElmFraCurYua:C17}. The yellow line at the center of the nadir is an artifact of the sensor itself and indicates where the port and starboard sonar returns are combined into an image. 

\subsection{Measurement Model}
\label{sec:MeasModel}

In this work, we assume landmark positions are available in a global frame of reference and that the seafloor depth does not vary in the cross-track direction. Each landmark with index $d \rmv\in\rmv \{1,\dots,D\}$ is represented by a rectangle described by the vector $\V m_d = [ x_d \ist\ist\ist\ist\ist y_d \ist\ist\ist\ist\ist  \theta_d \ist\ist\ist\ist\ist  l_d \ist\ist\ist\ist\ist w_d ]^{\T}\rmv\rmv\rmv$. The 2-D Cartesian coordinates of the center of the landmark are denoted by $[x_d \ist\ist\ist y_d]^{\T}$. $\theta_d$ is the landmark orientation in radians, where $\theta_d = 0$ when the vehicle points directly east and $\frac{\pi}{2}$ when the vehicle points north. $[l_d \ist\ist\ist w_d]^{\T}$ denotes the respective length and width of the landmark, which we are assuming to be nearly flat on the seafloor. For this reason, we do not consider landmark height at this time.

At time $k$, a new landmark detection is defined as $\V{z}_{k,l} = \big[{z}^{(1)}_{k,l} \ist\ist\ist\ist {z}^{(2)}_{k,l} \big]^{\T}\rmv\rmv\rmv$, $l \in \{1,\dots,L_k\}$, where $L_k$ is the total number of detections. The two elements of the detection vector are noisy slant range to the near and far edge of the landmark. Negative and positive distances denote the range to the port and starboard side of the vehicle track respectively. Here we describe the model that relates a \ac{sss} measurement (i.e., a ping) to the vehicle state. This model will be used in the navigation filter discussed in Section~\ref{sec:Methods}. 

At time $k$, the \ac{sss} ping covers a line segment on the seafloor that has port and starboard end points defined\vspace{3mm} by
\begin{align}
\V x^{(\mathrm{p})}_{k} &= \begin{bmatrix}
x_k\\
y_k\\
\end{bmatrix}
+
\begin{bmatrix}
\cos(\theta_k + \frac{\pi}{2})\\
\sin(\theta_k + \frac{\pi}{2})\\
\end{bmatrix}
\sqrt{r_{\mathrm{max}}^2 - \gamma_k^2} \nn\\[3.5mm]
\V x^{(\mathrm{s})}_{k} &= 
\begin{bmatrix}
x_k\\
y_k\\
\end{bmatrix}
- 
\begin{bmatrix}
\cos(\theta_k + \frac{\pi}{2})\\
\sin(\theta_k + \frac{\pi}{2})\\
\end{bmatrix}
\sqrt{r_{\mathrm{max}}^2 - \gamma_k^2} \label{eq:q1} 
\vspace{1mm}
\end{align}
where $r_{\mathrm{max}}$ is the maximum slant range of the sensor. A box giving the globally referenced location of each landmark $d$ is defined by line segments with the following beginning and end\vspace{1mm} points:
\begin{align}
\V x^{\mathrm{b}}_{d,s}
&= \begin{bmatrix}
x_{d}\\
y_{d}\\
\end{bmatrix}
+ b^{(1)}_{s} \ist \frac{1}{2} \rmv\rmv
\begin{bmatrix}
l_d(\sin(\theta_d) - \cos(\theta_d)) \\
-w_d(\sin(\theta_d) + \cos(\theta_d))\\
\end{bmatrix} \nn \\[3mm]
\V x^{\mathrm{e}}_{d,s} &=
\begin{bmatrix}
x_{d}\\
y_{d}\\
\end{bmatrix}
+ b^{(2)}_{s} \ist \frac{1}{2} \rmv\rmv
\begin{bmatrix}
l_d(\sin(\theta_d) + \cos(\theta_d))\\
w_d(\sin(\theta_d) - \cos(\theta_d))\\
\end{bmatrix}
\label{eq:l2}
\vspace{1mm}
\end{align}
where $\V{b}_{s} = \big[b^{(1)}_{s} \ist\ist b^{(2)}_{s}\big]^{\T}\rmv\rmv\rmv$, $s \rmv\in\rmv \{1,2,3,4\}$ is given\vspace{.3mm} by $\V{b}_{1} = [1 \ist\ist\ist 1]^{\T}\rmv\rmv$, $\V{b}_{2} = [1 \ist\ist\ist \rmv\rmv-\rmv\rmv\rmv1]^{\T}$\rmv\rmv, $\V{b}_{3} = [-\rmv1 \ist\ist\ist \rmv\rmv-\rmv\rmv\rmv1]^{\T}$\rmv\rmv, and $\V{b}_{4} = [-\rmv1 \ist\ist\ist 1]^{\T}\rmv\rmv$.

Let $h_1(\V m_d,  \V{x}_k)$ be the binary function that is equal to $1$ if the line segments defined in \eqref{eq:q1} intersect with the line segments defined in \eqref{eq:l2}, and 0 otherwise. Specifically, $h_1(\V m_d,  \V{x}_k) = 1$ when the landmark with index $d$ is present in the line on the seafloor covered by the SSS ping. If landmark $d$ is not present, then $h_1(\V m_d,  \V{x}_k) = 0$. The intersection of these line segments is necessary for the landmark to generate a detection. Due to sensor noise, however, it is not a sufficient condition. In other words, physically present landmarks generate detections with probability $0 \rmv<\rmv P_{\mathrm{det}} \rmv\leq\rmv 1$, which we assume to be known. Given the binary function $h_1(\V m_d,  \V{x}_k)$, the probability that a landmark generates a detection is described by a detection probability function $0\leq p_{\mathrm{det}}(\V m_d,  \V{x}_k) \leq 1$, which is equal to $P_{\mathrm{det}}$ for $h_1(\V m_d,  \V{x}_k) = 1$ and $0$ for $h_1(\V m_d,  \V{x}_k) = 0$. 

Whenever there is an SSS detection, there will be two intersection points, one representing each side of the landmark. When a landmark is identified on the edge of the ping, the second landmark intersection will be at $r_{\mathrm{max}}$. For the case $h_1(\V m_d,  \V{x}_k) \rmv=\rmv 1$, let $h_2(\V m_d,  \V{x}_k)$ be the function that provides the intersections of lines \eqref{eq:q1} and \eqref{eq:l2} in the form of $\big[\V x^{(1)}_{k,d} \ist\ist\ist \V x^{(2)}_{k,d} \big]^{\T}\rmv\rmv\rmv\rmv$. For a detection generated by landmark $d$, the measurement model is given by $\V{z}_{k,d} \rmv\rmv= h_2(\V x_k,\V m_d) + \V n_{k,d}$, where $\V n_{k,d}$ is zero-mean Gaussian noise with covariance matrix $\sigma^2_d \ist \V I_2$. The noise is assumed to be statistically independent across time and individual detections. The function $\big[r^{(1)}_{k,d} \ist\ist\ist r^{(2)}_{k,d} \big]^{\T} \rmv\rmv\rmv= h_3(\V x^{(1)}_{k,d}, \V x^{(2)}_{k,d},\V{x}_k)$ provides the distances of the two intersection points with respect to vehicle state $\V{x}_k$ in slant range,\vspace{.5mm} i.e.,
\begin{align}
\begin{bmatrix}
       r^{(1)}_{k,d}  \\[1mm]
       r^{(2)}_{k,d}
\end{bmatrix} \rmv\rmv=\rmv\rmv \begin{bmatrix}
       \sqrt{\big\| \V x^{(1)}_{k,d} - [x_k \ist\ist\ist\ist y_k ]^{\T} \big\|^2 + \gamma^2_k} \phantom{,}  \\[1em]
      \sqrt{\big\| \V x^{(2)}_{k,d} - [x_k \ist\ist\ist\ist y_k ]^{\T} \big\|^2 + \gamma^2_k} \phantom{,} \phantom{,}\\[0.3em]
     \end{bmatrix}\rmv\rmv. \label{eq:measurementModel}
     \vspace{1mm}
\end{align}
This gives a final measurement model $h(\V x_k,\V m_d,\V n_{k,d}) = h_3(h_2(\V m_d,  \V{x}_k),\V{x}_k)$. In summary, $h_1(\V m_d,  \V{x}_k)$ indicates the physical presence of a landmark, $p_{\mathrm{det}}(\V m_d,  \V{x}_k)$ models the probability that that landmark generates a detection, $h_2(\V m_d,  \V{x}_k)$ provides the points of intersection of the landmark edges with the sonar ping, and $h_3(\V x^{(1)}_{k,d}, \V x^{(2)}_{k,d},\V{x}_k)$ provides slant ranges to the landmark based on the output from $h_2(\V m_d,  \V{x}_k)$.

We define the vector of \ac{sss} detections at time $k$ as $\V z_k = [\V z_{k,1},\dots, \V z_{k,L_k}]^{\T}\rmv$. If $\V z_{k,l}$ was generated by landmark $d$, the likelihood function corresponding to this measurement model is given by
\begin{align}
p(\V z_{k,l}|\V x_k; \V m_d) = 
\mathcal{N}\big(h_3(h_2(\V m_d,  \V{x}_k),\V{x}_k) ;\sigma^2_d \M{I}_2 \big). \label{eq:trueFalse}
\vspace{1mm}
\end{align}
However, due to \ac{mou} the index $l$ does not provide any information on the index $d \in \{1 \dots D\}$. This is driven by uncertainty in the vehicle state and the presence of false positives, which follow a known clutter \ac{pdf} $f_{\mathrm{c}}(\V z_{k,l})$. In addition to sonar detections, vehicle height above the seafloor and heading are modeled as $y_{k,\mathrm{h}} =  \gamma_k + n_{k,\mathrm{h}}$ and $y_{k,\mathrm{c}} =  \theta_k + n_{k,\mathrm{c}}$. $n_{k,\mathrm{h}}$ and $n_{k,\mathrm{c}}$ are zero-mean statistically independent noise terms with variances $\sigma_{\mathrm{h}}^2$ and $\sigma_{\mathrm{c}}^2$. These noise terms are also statistically independent of landmark detection noise ${n}_{k,d}$. The total measurement vector of length $2L_k+2$ is given by $\V{y}_{k} = [\V{z}_{k,1},  \dots, \V{z}_{k,L_k},  \ist\ist\ist y_{k,\mathrm{h}}, \ist\ist\ist y_{k,\mathrm{c}} ]^{\T} $. The joint likelihood function used in the navigation filter developed in Section~\ref{sec:Methods} can be expressed as
%\begin{equation}
%p(\V{y}_k | \V{x}_k) = p(y_{k,\mathrm{c}} | \V{x}_k) \ist p(y_{k,\mathrm{h}} | \V{x}_k)  \prod^{D}_{d=1} p(\V{z}_{k,d} | \V{x}_k) \label{eq:jointLikelihood}
%\end{equation}
\begin{equation}
p(\V{y}_k | \V{x}_k) = \tilde{p}(\V{z}_k, L_k | \V{x}_k) \ist p(y_{k,\mathrm{c}} | \V{x}_k) \ist p(y_{k,\mathrm{h}} | \V{x}_k).  \label{eq:jointLikelihood}
\end{equation}
Here, $\tilde{p}(\V{z}_k, L_k | \V{x}_k)$ is an approximation of the likelihood function for all landmark detections that takes into account \ac{mou}, the details of which are derived in the following section.

\section{MOU Model for SSS Measurements}
\label{sec:measurementsAndMOU}

In general, it is unknown which landmark corresponds to each sonar detection. In addition, sensor noise leads to false positives and certain landmarks, despite being in the field of view of the sensors, may not generate a detection (``missed detection''). In what follows, we develop a statistical framework inspired by joint \ac{pda}. \cite{BarWilTia:B11,MeyKroWilLauHlaBraWin:J18}, that explicitly addresses \ac{mou} with \ac{sss} detections. Our model statistically describes sensor noise as well as noise due to the landmark detection extraction process. As in joint \ac{pda}, a key aspect of our model is to represent the uncertain measurement-to-landmark associations by a random association vector, $\V a_k = [a_{1,k}, \dots a_{D,k}]$ with $a_{d,k} \in \{0 \dots L_k\}$. This association vector identifies the origin of each detection by assigning elements of the detection vector $\V z_k = [\V z_{k,1},\dots, \V z_{k,L_k}]$ to landmark indices \cite{BarDauHua:J09, BarWilTia:B11}. $a_{d,k} \rmv=\rmv l$ indicates that landmark $d$ has generated detection $l$, whereas $a_{d,k} \rmv=\rmv 0$ indicates that landmark $d$ has not generated a detection at time $k$.

We also make the following assumptions that are common to handling \ac{mou}: (i) each landmark detection follows an independent Bernoulli trial with probability, $p_{\mathrm{det}}(\V m_d,  \V{x}_k)$, (ii) the number of clutter detections are Poisson distributed with mean $\mu_{\mathrm{c}}$, (iii) each landmark generates at most one detection, and (iv) each detection is generated by at most one landmark. In other words, two landmarks cannot generate the same detection, and a single landmark cannot generate more than one detection \cite{KscFreLoe:01}. Condition (iii) is automatically satisfied by the definition of association vector $\V{a}_k$ and (iv) is enforced by a check function, $\varphi (\V a_k)$, which is zero if condition (iv) is violated, i.e., if two elements of  the association vector $\V a_k$ are the same \cite{MeyKroWilLauHlaBraWin:J18, MeyWil:J21}. Based on the assumptions stated above, the \ac{pmf} of $\V a_k$ and $L_k$ given the vehicle state $\V x_k$ can be derived as follows \cite{BarWilTia:B11,MeyKroWilLauHlaBraWin:J18}, \vspace{.5mm}
\begin{equation}  % add semicolon Md because pdet is fxn of m and x
p(\V a_k, L_k | \V{x}_k) = \varphi (\V a_k) \biggl( \hspace{.3mm} \prod_{d \in \mathcal{D}_{\V a_k} } \frac{p_{\mathrm{det}}(\V m_d,\V{x}_k)}{\mu_{\mathrm{c}}(1-p_{\mathrm{det}}(\V m_d,\V{x}_k))}\biggr) \frac{\mathrm{e}^{-\mu_{\mathrm{c}}}\mu_{\mathrm{c}}^{L_k}}{L_k!} \biggl(1-p_{\mathrm{det}}(\V m_d,\V{x}_k)\biggr)^{\rmv\rmv D}. \label{eq:general_pak}
\vspace{1mm}
\end{equation}
$\mathcal{D}_{\V a_k} = \{ d \in \{1 \dots D\}| a_{d,k} > 0\}$ is the set of landmarks that, according to vector $\V a_k$, generated a positive detection. Given that $f_{\mathrm{c}}(\V z_{k,l})$ is the \ac{pmf} of the \ac{iid} clutter detections, the likelihood of the \ac{sss} measurement vector at time $k$ can be expressed as \vspace{0mm}\cite{BarWilTia:B11,MeyKroWilLauHlaBraWin:J18}
\begin{equation}
p(\V{z}_k | \V{x}_k,  \V{a}_k, L_k) = \Bigg( \prod^{D}_{d = 1} q_1(\V x_k, a_{d,k}; \V{z}_k, \V m_d) \Bigg) \prod^{L_k}_{l = 1} f_{\mathrm{c}}(\V z_{k,l})\label{eq:doubleProduct}
\vspace{1mm}
\end{equation}
where the function $q_1(\V x_k,a_{d,k}; \V{z}_k, \V m_d)$ is given by
\begin{align}
q_1(\V x_k,a_{d,k}; \V{z}_k, \V{m}_d) = \begin{cases}
    \frac{p(\V z_{a_{d,k}}|\V x_k, \V m_d)}{f_{\mathrm{c}}(\V z_{a_{d,k}})} & a_{d,k} \in \{1,\dots,L_k\} \\
    1 & a_{d,k} = 0.
\end{cases}\label{eq:qfunc1}
\vspace{1mm}
\end{align}
In the case where all detections are clutter,  $\V{a}_k$ is all zeros and all factors $q_1(\V x_k,a_{d,k}; \V{z}_k, \V{m}_d) $, $d \in \{1,\dots,D\}$ are equal to $1$. In that case, \eqref{eq:doubleProduct} reduces to a product of clutter \acp{pdf}. For each entry $a_{d,k} > 0$, the numerator of $q_1(\V x_k,a_{d,k}; \V{z}_k, \V{m}_d)$ represents a detection-to-landmark association, while the denominator cancels a factor of $f_{\mathrm{c}}(\V z_{a_{d,k}})$ in \eqref{eq:doubleProduct}. 

For the derivation of our sequential Bayesian estimation method, the \acp{pdf} in  \eqref{eq:general_pak}  and \eqref{eq:doubleProduct} can be further simplified by the following: (i) at each time step the detections $\V z_k$ (and thus $L_k$) are fixed, (ii) to evaluate the likelihood function in our filter update we only need to know \eqref{eq:jointLikelihood} up to a normalization constant. For fixed $L_k$, we can simplify and rewrite the \ac{pdf} in \eqref{eq:general_pak}\vspace{-1.5mm} as
\begin{align}  % add semicolon Md because pdet is fxn of m and x
p(\V a_k, L_k | \V{x}_k) &\propto \varphi (\V a_k) \prod^{D}_{d = 1} q_2(\V x_k,a_{d,k}; \V{m}_d) \label{eq:pmf}
\vspace{1mm}
\end{align}
where the function $q_2(\V x_k,a_{d,k}; \V{m}_d)$ is defined\vspace{.5mm} as 
\begin{equation}
q_2(\V x_k,a_{d,k}; \V{m}_d) = \begin{cases}
    \frac{p_{\mathrm{det}}(\V{m}_d,\V{x}_k)}{\mu_{\mathrm{c}}} & a_{d,k} \in \{1,\dots,L_k\} \\
    1-p_{\mathrm{det}}(\V m_d,\V{x}_k) & a_{d,k} = 0.
\end{cases}\label{eq:qfunc2}
\vspace{1mm}
\end{equation}
Note that in \eqref{eq:pmf}, we use the symbol ``$\propto$'' to indicate equality up to a constant factor because we dropped the constant $\mathrm{e}^{-\mu_{\mathrm{c}}}\mu_{\mathrm{c}}^{L_k}/L_k!$. Next, we combine equations \eqref{eq:general_pak} and \eqref{eq:doubleProduct} using the chain rule of \acp{pdf} and ``marginalizing out'' $\V a_k$, i.e.,\vspace{1mm}  
\begin{align}
p(\V{z}_k, \V{L}_k | \V{x}_k) =  \sum_{\V{a}_k} p(\V{z}_k, \V{L}_k , \V{a}_k  | \V{x}_k)  \nn\\
=  \sum_{\V{a}_k} \ist  p(\V{z}_k | \V{x}_k,  \V{a}_k, L_k) \ist p(\V{a}_k, L_k|\V{x}_k).  \label{eq:likelihoodSonarData}
\vspace{1mm}
\end{align}
Using expression \eqref{eq:doubleProduct} for  $p(\V{z}_k | \V{x}_k,  \V{a}_k, L_k)$ and expression \eqref{eq:pmf} for $p(\V{a}_k, L_k|\V{x}_k)$, expression \eqref{eq:likelihoodSonarData} can be rewritten\vspace{1mm} as 
\begin{align}
\hspace{-.5mm}p(\V{z}_k, \V{L}_k | \V{x}_k)  &\propto  \sum_{\V{a}_k} \varphi (\V a_k) \prod^{D}_{d=1} g(\V x_k, a_{d,k}; \V{z}_k , \V{m}_d )
\label{eq:likelihoodFunction}
\end{align}
where we introduced $g(\V x_k, a_{d,k}; \V{z}_k , \V{m}_d ) = q_1(\V x_k, a_{d,k};\V{z}_k,$ $\V{m}_d )  \ist q_2(\V x_k, a_{d,k}; \V{m}_d ).$ Note that here we have dropped the second product in \eqref{eq:doubleProduct} because for fixed observations that product is a constant.

Finally, to address the exponential computational complexity related to the summation of all possible association events, i.e., the summation $\sum_{\V{a}_k}$, we perform the belief propagation approximation of the consistency constraint $\varphi (\V a_k)$ discussed in \cite{MeyKroWilLauHlaBraWin:J18}. Specifically, we perform an accurate approximation of the form $\varphi (\V a_k) \approx \prod^{D}_{d=1} \kappa(a_{d,k})$ (see \cite[Sec.~VI and VII]{MeyKroWilLauHlaBraWin:J18} for details on how to compute $\kappa(a_{d,k})$, $d \in\{1,\dots D\}$). 

%Note that $\kappa(a_{d,k} = 0) = 1$ for $d \in \{1,\dots,D\}$.%

Based on this approximation, we further simplify \eqref{eq:likelihoodFunction} by changing the order of the summation and the product. Given the aforementioned approximation of $\varphi (\V a_k)$, an approximation of $p(\V{z}_k, \V{L}_k | \V{x}_k) $ is given by\vspace{-1mm}
\begin{align}
\tilde{p}(\V{z}_k, \V{L}_k | \V{x}_k) \propto  \prod^{D}_{d=1} \sum^{L_k}_{l = 0} \kappa(a_{d,k} = l) g (\V x_k, a_{d,k} = l; \V{z}_k, \V m_d ). 
\label{eq:BPSimplified} 
\vspace{1mm}
\end{align}

This approximation avoids the exponential increase of computational complexity in the number of landmarks related to the sum over all $\V{a}_k$. According to \eqref{eq:BPSimplified}, the joint likelihood function that includes all landmarks can be interpreted as a product of the individual likelihood functions where each consists of a weighted sum with $L_k + 1$ terms. Each term represents a possible detection-to-landmark association event. We simplify the evaluation of \eqref{eq:BPSimplified} by removing all factors related to landmarks outside the field of view, i.e., for which $p_{\mathrm{det}}(\V m_d,  \V{x}_k) = 0$. It can be easily verified that factors in \eqref{eq:BPSimplified} corresponding to these impossible-to-see landmarks \blk{are equal to a constant, due to $h_1(\V m_d,  \V{x}_k) = 0$ and thus $1\rmv-\rmv p_{\mathrm{det}}(\V m_d,\V{x}_k) = 1$  (cf.~definition of $g (\V x_k, a_{d,k};$ $\V{z}_k, \V m_d)$, \eqref{eq:qfunc1}, and \eqref{eq:qfunc2})}. Thus, without any further approximation, \eqref{eq:BPSimplified} can be simplified to 
\begin{align}
\tilde{p}(\V{z}_k, \V{L}_k | \V{x}_k)
\propto \rrmv\rrmv \prod_{d \in \mathcal{D}(\V{x}_k)} \sum^{L_k}_{l = 0} \kappa(a_{d,k} = l) g (\V x_k, a_{d,k} = l; \V{z}_k, \V m_d ). 
\label{eq:BPSimplified2}
\vspace{1mm}
\end{align}
where $\mathcal{D}(\V{x}_k) \subseteq \{1,\dots,D\}$ consists of all $d \in \{1,\dots,D\}$ with $h_1(\V m_d,  \V{x}_k) = 1$. It can also be verified (cf.~definition of $g (\V x_k, a_{d,k};$ $\V{z}_k, \V m_d)$, \eqref{eq:qfunc1}, and \eqref{eq:qfunc2}) that for the case $L_k = 0$, the expression in \eqref{eq:BPSimplified2}\vspace{-.5mm} reads
\begin{align}
\tilde{p}(\V{z}_k, \V{L}_k | \V{x}_k) \propto  \prod_{d \in \mathcal{D}(\V{x}_k)} \big ( 1-p_{\mathrm{det}}(\V m_d,\V{x}_k) \big). \label{eq:BPSimplified0}
\end{align}
Thus, even in the absence of any detections, $\tilde{p}(\V{z}_k, \V{L}_k | \V{x}_k)$ can provide information about the vehicle state $\V{x}_k$ because predicted vehicle states that incorrectly predict positive detections will be associated with a lower likelihood. \blk{The fact that the set $ \mathcal{D}(\V{x}_k)$ is a function of the $\V{x}_k$, makes a particle-based evaluation of \eqref{eq:BPSimplified0} tedious to implement efficiently since the number of considered landmarks can change across particles. Alternatively, $ \mathcal{D}(\V{x}_k)$ can be approximated by a set, $ \mathcal{D}_k$, that is independent of $\V{x}_k$, in a processing step known as gating \cite{BarShalom:B95}. In what follows, we refer to landmarks with indexes in $ \mathcal{D}_k$  as ``gated landmarks''.}

\section{The Navigation Filter}
\vspace{-.5mm}
\label{sec:Methods}

\begin{figure*}[!htb]
\begin{center}
  \includegraphics[width=1\linewidth]{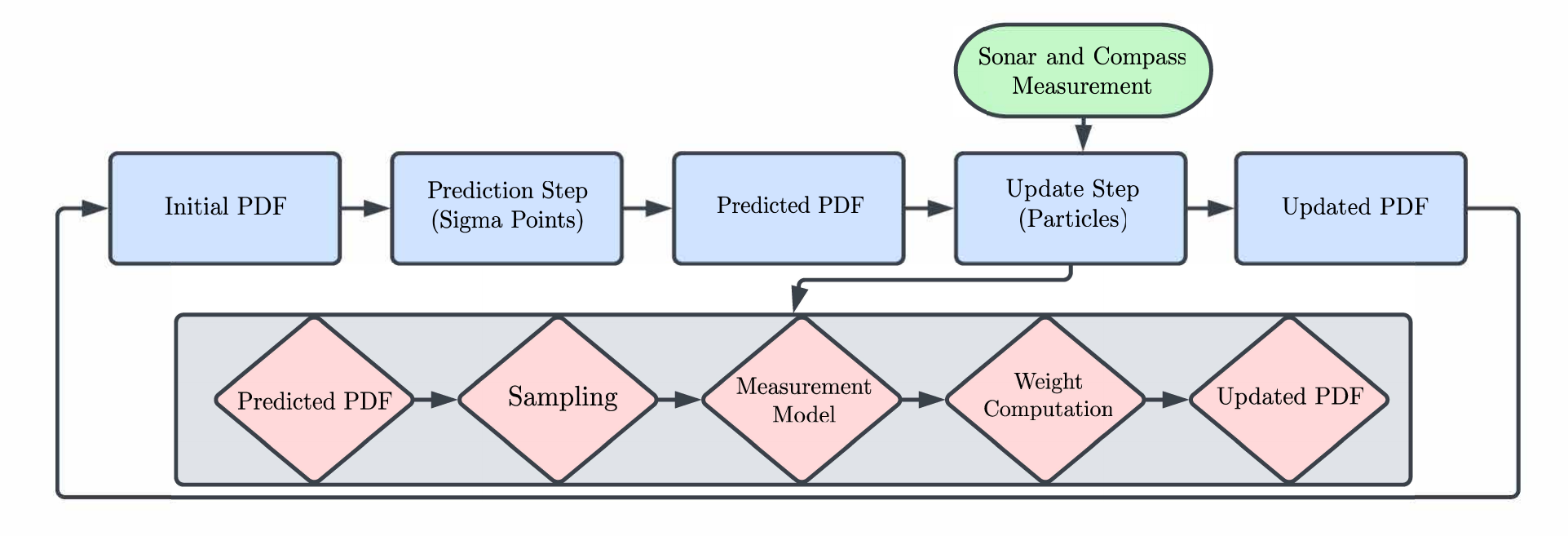}
  \vspace{-7mm}
\end{center}
  \caption{Flow diagram describing the order of operations in the navigation algorithm. Blocks shown in blue depict the main processing steps and their results. The sub-process in red shows details of the particle-based update step, which includes probabilistic association to handle \ac{mou}. The probabilistic association of landmarks and detections is performed in the weight computation portion of the update step.}
  \vspace{0mm}
  \label{fig:flowchart}
\end{figure*}

At time $k$, we aim to estimate the state, $\V{x}_{k}$, of the \ac{auv} from all measurements $\V{y}_{1:k}$. Given the conditional \ac{pdf} of the state, $p(\V x_k|\V y_{1:k})$, the \ac{mmse} estimate \cite{Kay:B93} of the state can be obtained\vspace{-1mm} as
\begin{equation}
   \hat{\V x}^{\mathrm{MMSE}}_{k} = \int \rmv \V x_k \ist p(\V x_{k}|\V y_{1:k}) \ist \mathrm{d}\V x_k. \label{eq:estimate}
\end{equation}
A Bayes filter\cite{AruMasGorCla:02} consisting of a prediction and update step is used to compute an approximation of the conditional \ac{pdf} $p(\V x_k|\V z_{1:k})$. The prediction step uses the Chapman-Kolmorogov equation applied to the state-transition model in \eqref{eq:stateTransitionModel}. The update step uses Baye's rule applied to the likelihood function in \eqref{eq:jointLikelihood}. We use sigma points \cite{JulUhl:04,WanMer:B01} in the prediction step and random samples, or ``particles'', \cite{AruMasGorCla:02} in the update step. We use the particles to compute the approximate posterior mean, $\V{\mu}_{k}$, and covariance matrix,  ${\V C_{k}}$, of $p(\V x_{k}|\V y_{1:k})$. Using particles rather than sigma points for the update step requires more computational resources but makes it possible to obtain accurate results despite a strongly nonlinear measurement model. In contrast to the measurement model, the nonlinearity in the state-transition model is moderate. A more efficient computation based on sigma points can thus provide accurate results and is preferred. Considering \eqref{eq:estimate}, the approximate posterior mean, $\V{\mu}_{k}$, computed by our filter is an approximation of the \ac{mmse} estimate, i.e., $\hat{\V x}^{\mathrm{MMSE}}_{k} \approx \V{\mu}_{k}$. The computation of $\V{\mu}_{k}$ handles \ac{mou} according to the methods described in \ref{sec:measurementsAndMOU}. The flow chart in Fig.~\ref{fig:flowchart} shows the navigation filter architecture. The statistical framework for prediction and update steps is described here.

\subsection{Prediction Step}\label{sec:MethodsSP}

Let, $\mathcal{N}(\V x_{k-1}; \V{\mu}_{k-1}, {\V C_{k-1}})$ be a Gaussian approximation of the marginal posterior \ac{pdf} $p(\V x_{k-1} | \V{y}_{1:k-1})$ computed at time step $k-1$. First, we introduce a state vector augmented by means of our driving noise terms, $\V{\eta}_{k-1} = [\V{\mu}^{\text T}_{k-1} \ist\ist\ist\ist\ist 0 \ist\ist\ist\ist\ist 0 \ist\ist\ist\ist\ist 0 \ist\ist\ist\ist\ist 0]^{\text T}$. In addition, we define a corresponding augmented covariance\vspace{-.5mm} matrix
\begin{equation}
\M{\Sigma}_{k-1} = \begin{bmatrix}
{\V C_{k-1}} &{\V 0}\\
{\V 0} &\mathrm{diag}({\V \sigma^2})\\
\end{bmatrix}
\vspace{.5mm}
\end{equation}
where $\mathrm{diag}({\V \sigma^2})$ denotes the diagonal matrix with diagonal elements given by driving noise variances $\V \sigma^2 = \begin{matrix} [\sigma^2_{\mathrm{s}} &\sigma^2_{\mathrm{t}} &\sigma^2_{\theta} &\sigma^2_{\gamma}\\ \end{matrix}]^{\text T}$. This augmentation makes the use of sigma points in the prediction step possible despite the nonlinear operation on $\V n_k$ in \eqref{eq:stateTransitionModel}. $2N+1$ sigma points are computed from the mean state according to \cite{JulUhl:04} and \cite{WanMer:B01}. In effect, $2 N$ sigma points are evenly spaced around one sigma point at $\V{\eta}_{k-1}$. The augmented sigma points, $\V{x}^{i}_{k-1}$, are defined\vspace{.5mm} as 
\begin{align}
    \V{x}^{i}_{k-1} &= \V{\eta}_{k-1}  + (\sqrt{N \M{\Sigma}_{k-1}})^{i} \quad i \in \{1,\dots,N\} \nn \\[2mm]
    \V{x}^{i}_{k-1} &= \V{\eta}_{k-1}  - (\sqrt{N \M{\Sigma}_{k-1}})^{i} \quad i \in \{N+1,\dots,2N\}. \nn
    \vspace{.8mm}
\end{align}
The weights corresponding to these sigma points are $w^{i} = \frac{1}{2 N}$ and the sigma point at $\V{\eta}_{k-1}$ has a weight of 0. Next, sigma points are passed through the state-transition model in \eqref{eq:stateTransitionModel}, \vspace{0mm} i.e.,
\begin{equation}\label{eq:vmm}
    \underline{\V x}^{i-}_{k} = g(\underline{\V{x}}^{i}_{k-1}, \overline{\V{x}}^{i}_{k-1}, \V{u}_k) \quad i \rmv\in\rmv \{1,\dots,2N\}. \nn
\end{equation}
Here, $\underline{\V{x}}^{i}_{k-1}$ and $\overline{\V{x}}^{i}_{k-1}$ denote the first four and last four elements of sigma point $\V x^{i}_{k}$. The first four elements correspond to a state vector; the last four are driving noise terms. The updated mean and covariance of the sigma points are computed according to
\begin{align} 
    \V{\mu}_k^{-} &= \sum_{i=1}^{2N}w^{i}{\underline{\V x}}^{i-}_{k} \nn \\[2mm]
    {\V C^{-}_{k}} &= \sum_{i=1}^{2N}w^{i}({\underline{\V x}}^{i-}_{k} -  \V{\mu}_k^{-})({\underline{\V x}}^{i-}_{k} -  \V{\mu}_k^{-})^{\text T}. \nn
\end{align}
These terms describe the predicted posterior \ac{pdf} $p(\V x_k | \V{y}_{1:k-1}) \approx\vspace{3mm} \mathcal{N}(\V x_k;  \V{\mu}_k^{-}  , {\V C^{-}_{k}})$.

\subsection{Update Step}\label{sec:MethodsPF}
\vspace{-.5mm}

We use a particle filter to compute the updated posterior \ac{pdf} $p(\V x_k | \V{y}_{1:k}) \rmv\approx\rmv \mathcal{N}(\V x_k; \V{\mu}_{k} , {\V C_{k}})$. Importance sampling is performed using the Gaussian representation $\mathcal{N}(\V x_k; \V \mu^{-}_{k} , {\V C^{-}_{k}})$ of the predicted posterior \ac{pdf} as a proposal \ac{pdf} \cite{HliHlaDju:13,AruMasGorCla:02}. According to these assumptions, we sample $I$ particles denoted as $\big\{\Bar{\V x}^{i}_{k}  \big\}^{I}_{i=1}$ from $\mathcal{N}(\V x_k; \V \mu^{-}_{k} , {\V C^{-}_{k}})$. Corresponding\vspace{-.5mm} particle weights, $\big\{\Bar{\V w}^{i}_{k}  \big\}^{I}_{i=1}$, are computed based on the joint likelihood provided in \eqref{eq:jointLikelihood},  which relies on the approximate likelihood function for \ac{sss} detections $\tilde{p}(\V{z}_k, \V{L}_k | \V{x}_k)$ in \eqref{eq:BPSimplified2}. 

To guarantee that $\tilde{p}(\V{z}_k, \V{L}_k | \V{x}_k)$ has the same number of factors for each particle, we replace the set $\mathcal{D}(\V{x}_k)$ in \eqref{eq:BPSimplified2} with an approximate set of feasible landmarks, $\tilde{\mathcal{D}}$. This subset is selected by gating that removes landmarks beyond the sensing distance of the \ac{sss}\cite{BarShalom:B95}. We assume that the true vehicle state is distributed according to the predicted posterior \ac{pdf} $p(\V x_k | \V{z}_{1:k-1}) \approx \mathcal{N}(\V x_k;  \V{\mu}_k^{-}, {\V C^{-}_{k}})$. The corresponding elliptically-shaped validation region is defined by\vspace{-1mm}
% x_{k+1} is the predicted state after vmm - udpate this notation
\begin{align}
V(k,\Gamma) = \big\{\V m_d: [{\V m_d} - {\V \mu}^{-}_{k}]^{\mathrm{T}}{\V C^{-}_{k}}^{-1}[{\V m_d}-{\V \mu}^{-}_{k}] \leq \Gamma \big\} \label{eq:validation}
\vspace{1mm}
\end{align}
where $\Gamma$ is a tunable constant. We choose $\Gamma = 6.6$, which corresponds to a probability of $0.99$. The set $\tilde{\mathcal{D}}_k$ is defined as the indexes of all landmarks inside this validation region. Using $\tilde{\mathcal{D}}_k$ and plugging the resulting from of \eqref{eq:BPSimplified2} into \eqref{eq:jointLikelihood}, we obtain the following expression for unnormalized particle weights,
\begin{align}
\tilde{w}_{k,d}^{i} = p(y_{k,\mathrm{c}} | \bar{\V{x}}^{i}_k) \ist p(y_{k,\mathrm{h}} | \bar{\V{x}}^{i}_k)
\times \prod_{d \in \tilde{\mathcal{D}}_k} \sum^{L_k}_{l = 0} \kappa(a_{d,k} = l) g (\bar{\V{x}}^{i}_k, a_{d,k} = l; \V{z}_k, \V m_d ) .  \label{eq:particleWeights}
\vspace{1mm}
\end{align}
Weights are computed and normalized in the log domain for numerical stability. After conversion to the linear domain, the final set of weights $\big\{ ( \V{x}^{i}_{k}, \Bar{w}^{i}_{k} ) \vspace{.3mm} \big\}^{I}_{i=1}$ is obtained from another normalization step s.t. $\sum^{I}_{i=1} \Bar{w}^{i}_{k} = 1$.  Finally, we compute the updated state estimate and covariance matrix below \cite{HliHlaDju:13, AruMasGorCla:02},
\begin{align}\label{eq:pfMean}
    \V \mu_{k} &= \sum_{i=1}^{I} \rmv \bar{w}_k^{i} \ist {\V x}^{i}_{k} \nn \\[1mm]
    {\V C_{k}} &= \sum_{i=1}^{I}\bar{w}^{i}({\V x}^{i}_{k} - \Bar{\V x}_{k})({\V x}^{i}_{k} - \Bar{\V x}_{k})^{\T}. \nn
\end{align}
This mean and covariance matrix are a Gaussian approximation of the updated marginal posterior \ac{pdf}, $p(\V x_{k} | \V{y}_{1:k}) \rmv= \mathcal{N}(\V x_{k}; \V \mu_{k}, {\V C_{k}})$. The posterior \ac{pdf} is then used in the prediction step at time $k+1$. To avoid particle degeneracy, a resampling step is performed\vspace{1mm} \cite{HliHlaDju:13, AruMasGorCla:02}.

\subsection{\blk{Scalability}}
\label{sec:complexity}

\blk{Due to the use of a large number of particles, the update step of the proposed method has a computational complexity that is significantly higher than the prediction step. We introduced two approximations to limit the computational complexity of the update step. In the first step, we use loopy belief propagation to efficiently approximate Equation \eqref{eq:likelihoodFunction} by Equation \eqref{eq:BPSimplified}. After this first approximation, the computational complexity of the update step at time $k$, only scales as $O(D L_k)$, i.e., proportional to the product of the number of landmarks $D$ and the number of measurements $L_k$. (Note that the original expression in Equation  \eqref{eq:likelihoodFunction} yields a computational complexity that scales exponentially in $D$ and $L_k$.) The second approximation consists of a gating step described by Equation \eqref{eq:validation}. Only landmarks within a reasonable distance of the vehicle are considered for data association and particle weight computations. This implies that the computational cost of Equation \eqref{eq:BPSimplified} does not depend on the total number of known landmarks but on the density of landmarks close to the vehicle. With these simplifications, the computational complexity is further reduced and only scales as $O(D_k \ist L_k)$ where $D_k \leq D$ is the number of gated landmarks. In the worst case, the number of measurements, $L_k$, increases linearly with $D_k$. Thus, the computation complexity of the update step has a worst-case scalability that is quadratic in the number of gated landmarks. Because of the uncertainty in vehicle location, the maximum range used for gating landmarks is significantly larger than the range of the sonar, and it is far more likely that $L_k$ increases slower than linearly with $D_k$. Thus, the complexity of the update step can be expected to scale slower than quadratically in the number of landmarks $D_k$. Section VI, further discusses the impact of gating and the number of landmarks on algorithm run time.}

\section{Simulation, Data Collection, and Processing}
\label{sec:DataCollect}

\begin{figure*}[!htb]
    \centering
    \subfloat[]{\resizebox{0.49\linewidth}{!}{  \includegraphics[scale=.52]{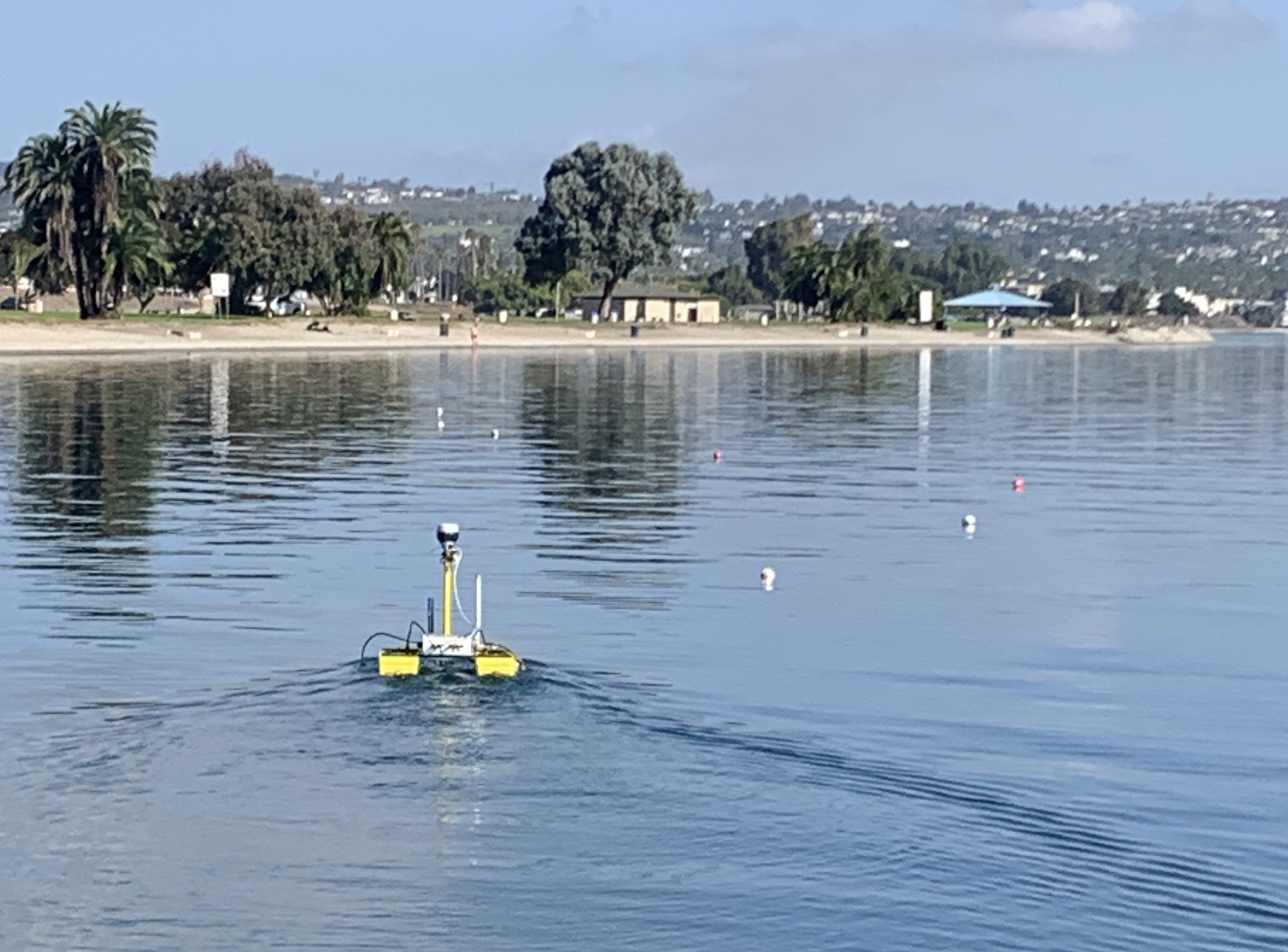} }} \hspace{1mm}
    \subfloat[]{\resizebox{0.48\linewidth}{!}{  \includegraphics[scale=.52]{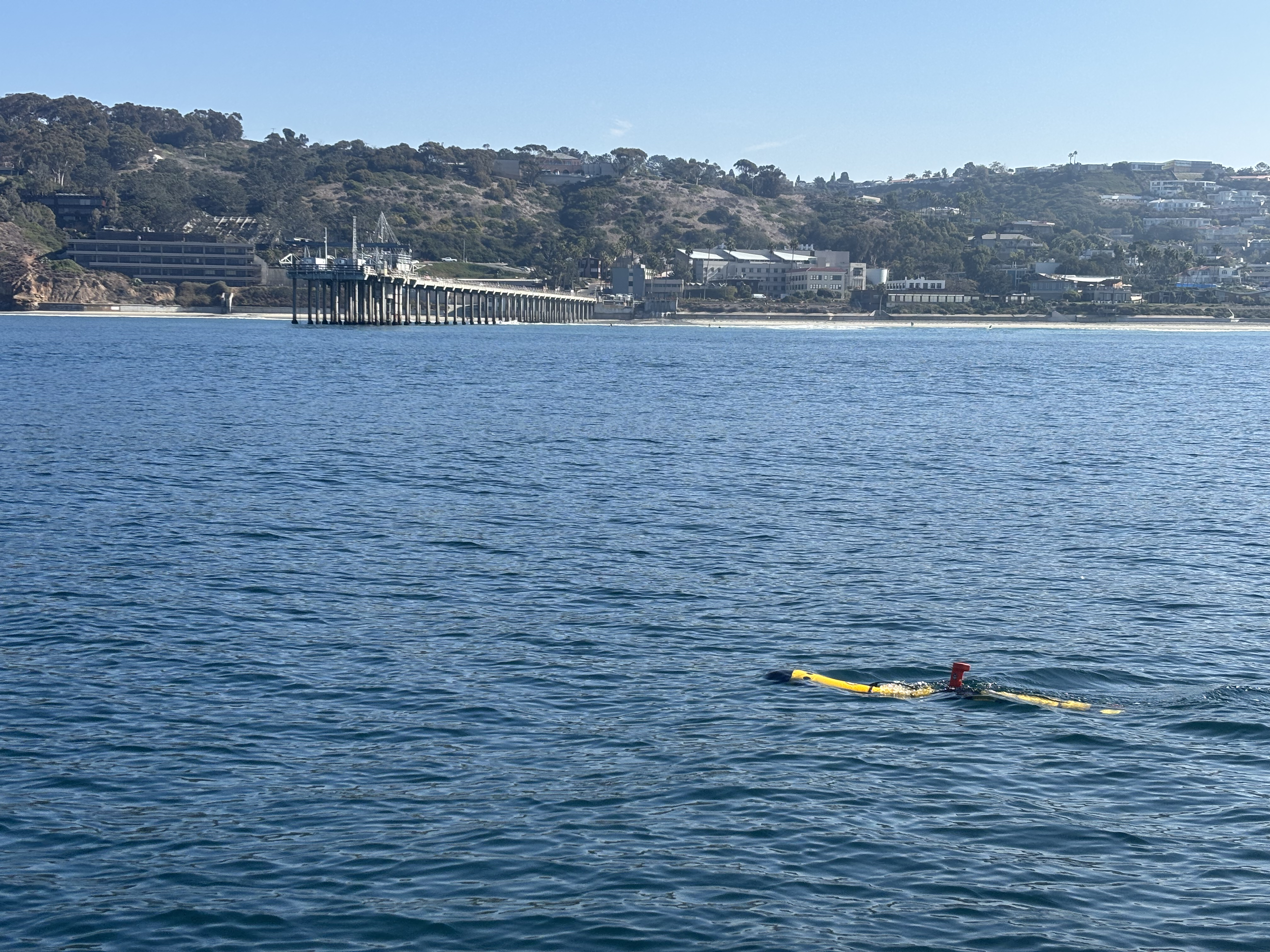} }}
    \caption{(a) \textit{Hydrone} surface vehicle with integrated \textit{Ark Scout Mk II} \ac{sss} collecting data in Mission Bay, CA. White and red buoys mark deployed landmarks. (b) \textit{Iver3} vehicle with \textit{EdgeTech} SSS collecting data at the Scripps Institution of Oceanography in La Jolla, CA. }\label{fig:hydrone}
    \vspace{-2mm}
\end{figure*}

We developed a forward simulation model that generates sonar data based on the measurement model from Section \ref{sec:MeasModel}. Similarly to related work \cite{StaBleUra:C08} and to our field experiment scenario, we assume a relatively flat sea bottom. We use the forward model for each simulated time step to compute the slant range to the near and far edges of landmarks in sight. We then add noise to these slant ranges. This process generates between 0 and $D$ \ac{sss} detections, depending on how many landmarks are present. We then add clutter to the landmark detections generated according to the Poisson \ac{pmf} described in Section \ref{sec:measurementsAndMOU}. This process results in a simulated observation vector of $L_k$ uncertain landmark detections. To bring the simulated scenario closer to the field experiment scenario, we add random displacements to the vehicle's location in the $x$ and $y$ directions. This emulates a persistent surface current and noisy wind-induced currents that are not directly incorporated in the state transition model. The mean of the current noise is 20 cm/s, with a standard deviation of 10 cm/s.

\blk{We performed two field experiments with two different vehicles and two different sites. Site 1 was Mission Bay, CA, where we used the Hydrone surface vehicle depicted in Fig. \ref{fig:hydrone}(a). The Hydrone was developed in collaboration with \textit{Seafloor Systems, Inc.} and is equipped with GPS, compass, and a \textit{Marine Sonic MkII} \ac{sss} sensor. The deployment took place in approximately 5 meters of water in an area with a flat, muddy bottom. Site 2 was La Jolla, CA, at the Scripps Institution of Oceanography where we deployed an Iver3 underwater vehicle made by \textit{L3Harris Technologies, Inc.} (Fig. \ref{fig:hydrone}(b)). The Iver3 platform is similarly equipped with GPS, compass, and an \textit{EdgeTech 2205} \ac{sss} sensor. It is a significantly more advanced system than the Hydrone and has several additional sensors, including a DVL. In contrast to the Hydrone, the Iver3 can dive which made is possible to operate in deeper water (approx. 6-9 meters). In our experiments, the vehicle flew approximately 5 meters above the seafloor. We used the proprietary Ivert3 vehicle localization method, which fuses compass, GPS, and DVL measurements as a ground truth. This site had a flat bottom and was predominantly sandy with occasional rocks. La Jolla experiments involved 6 landmarks, while experiments in Mission Bay involved 7 landmarks.}

We built artificial landmarks to be used in conjunction with two repurposed SeaSpider \ac{adcp} mounts \cite{SeaSpider}. The artificial landmarks were constructed using three sheets of high-density polyethylene (HDPE). The three sheets were zip-tied to form a triangular prism with two open ends. The open ends of the prism allow for easy deployment in the water as the landmark sinks quickly and can be retrieved with minimal effort. HDPE was chosen due to its low price, low weight, and reasonably high acoustic impedance \cite{Sel:J85}. \blk{The landmark configurations can be seen in Fig. \ref{fig:fieldTest}(a) and (b), where GPS coordinates have been converted into a local coordinate system. In Mission Bay (site 1), we performed three 7-minute deployments, driving the surface vehicle in a lawnmower pattern through the landmarks. In La Jolla (site 2), we performed three 11-minute deployments in which the vehicle autonomously navigated in a lawnmower pattern through the landmarks using waypoints.}

\section{Results}
\label{sec:Results}

Next, we report simulation and experimental results assessing the performance of our method.
\subsection{Simulation}

\begin{figure}[h]
\begin{center}
  \includegraphics[scale=.6]{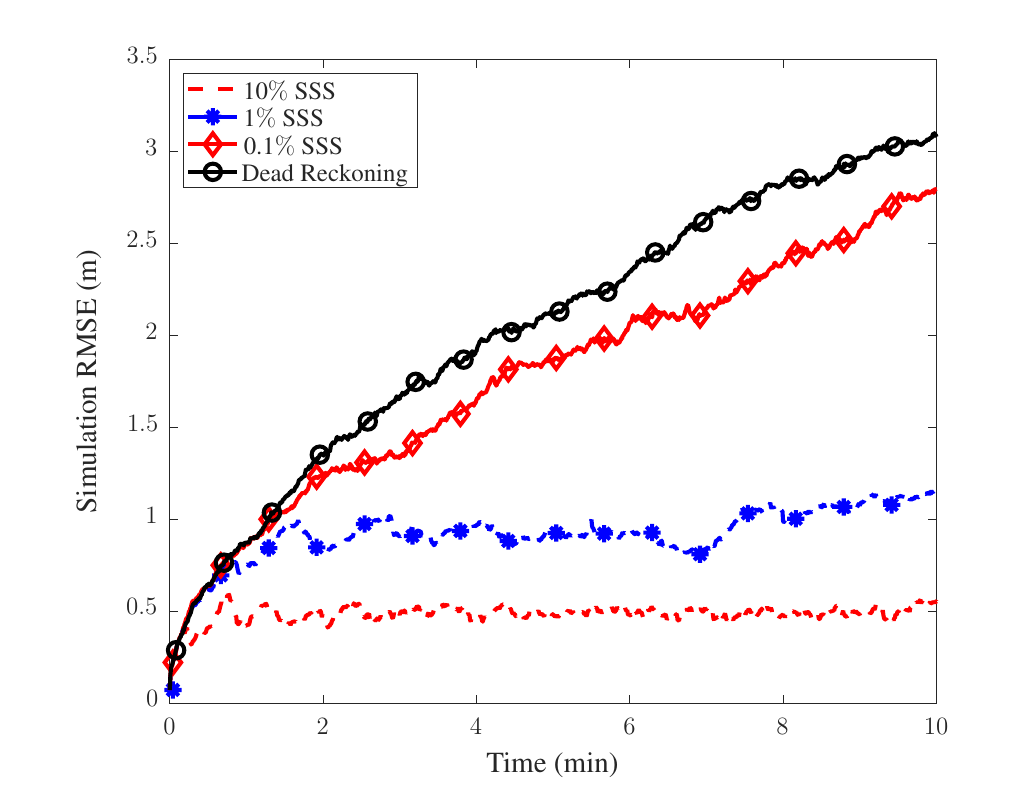}
\end{center}
  \caption{Average \ac{rmse} versus time for four different scenarios with the proposed navigation filter. The \ac{rmse} of the reference \ac{dr} method that does not use landmark information is shown in black. The other three curves correspond to scenarios where the likelihood of seeing a landmark is roughly 10$\%$, 1$\%$, and 0.1$\%$. \ac{rmse} curves for each scenario have been averaged over 300 simulation runs.}
  \label{fig:error}
\end{figure}

We assess simulation performance using \ac{rmse} of the vehicle position in 3 dimensions, the results of which can be seen in Fig. \ref{fig:error}. In each scenario, landmarks were located on an evenly-spaced grid, and the control input to the vehicle was random. There were five scenarios with landmark spacings ranging from 25 to 200 meters and one with no landmarks (i.e., dead reckoning). Results for each scenario were averaged over 300 simulation runs. 

Each simulation run has different random instances of driving and measurement noise. To generalize from set landmark spacing (e.g., 25 meters) to a real-world scenario, we evaluate the results in terms of the probability of observing a landmark. The results in Fig. \ref{fig:error} show algorithm performance given that the probability of observing a landmark is roughly 10$\%$, 1$\%$, and 0.1$\%$. These scenarios are compared to the performance of the \ac{dr} method (black). It can be seen that the \ac{rmse} of the location estimate provided by \ac{dr} grows unbounded as time evolves. On the other hand, when landmarks are observed frequently, the \ac{rmse} of the \blk{landmark-aided} localization method remains bounded. For the estimate without landmarks, the final \ac{rmse} of a 10-minute trajectory averaged 3.1 meters. When landmarks were observed 10$\%$ of the time, the estimation error was bounded around 0.5 meters. 

The key findings in the simulation are that (i) landmark spacing has an important effect on algorithm performance and (ii) the presence of landmarks can limit the position error. According to the simulation results, sightings must be frequent for the estimation error to remain bounded. For example, the vehicle location error is bounded when the probability of seeing a landmark is 1$\%$, but it is unbounded when that decreases to 0.1$\%$. Determining a minimum necessary frequency for landmark sightings will depend on system configuration and environmental parameters and remains a defining challenge of this method. In the field, there is a practical lower limit to the algorithm's accuracy, which is tied to the sensor resolution and environment noise. In general, we expect the \ac{rmse} in all scenarios to be higher in the field than in simulation.

\subsection{Field Experiments}

\begin{figure*}[!htb]
    \centering
    \subfloat[]{\resizebox{0.495\linewidth}{!}{  \includegraphics[scale=.7]{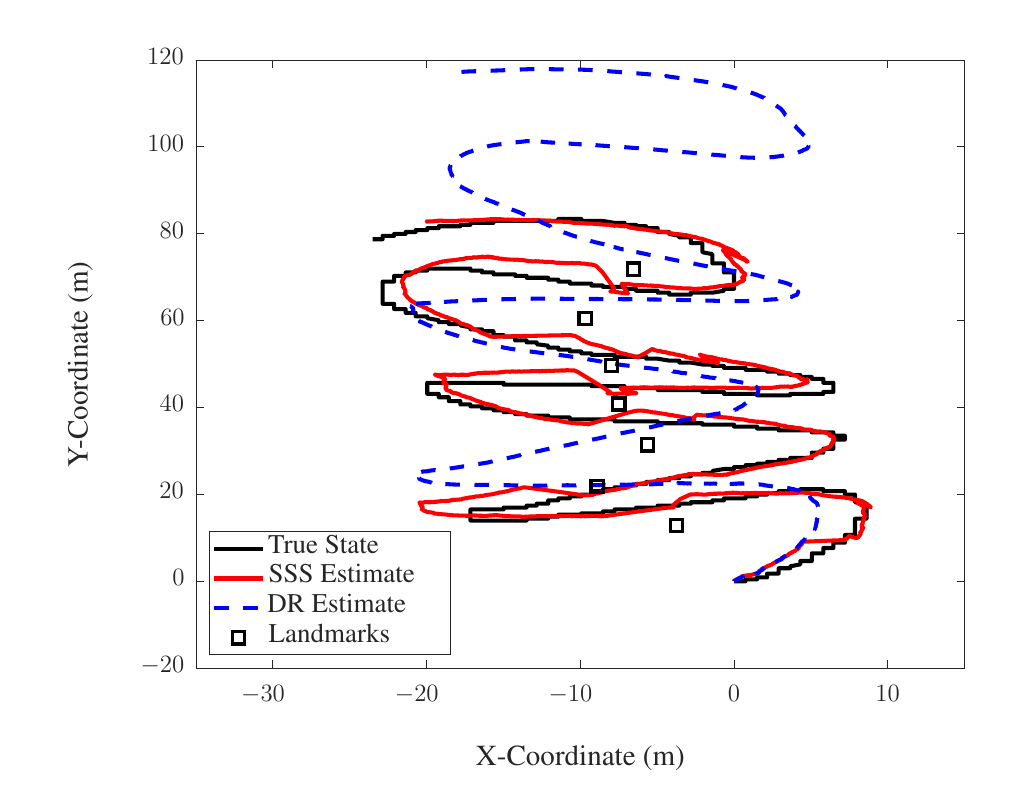} }} \hspace{1mm}
    \subfloat[]{\resizebox{0.49\linewidth}{!}{  \includegraphics[scale=.7]{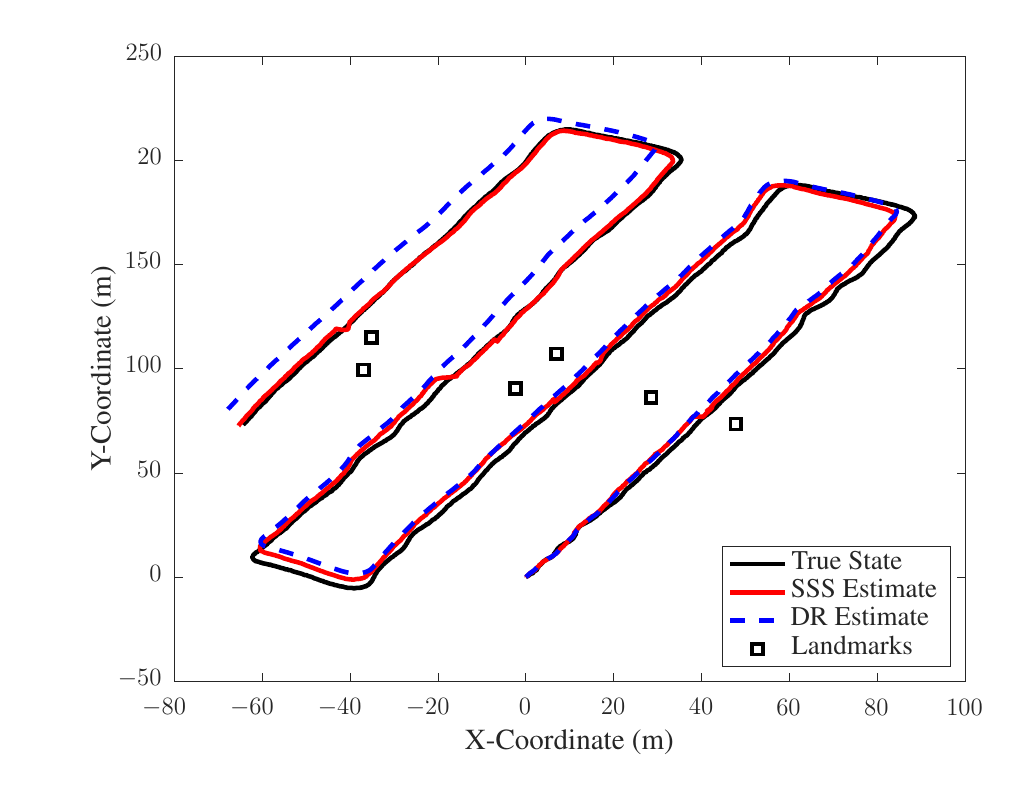} }}
    \caption{The three tracks in each figure show true vehicle location (black), estimated location with \ac{dr} (blue dashed), and estimated location with \ac{sss} (red). Artificial landmarks are shown in black. The Y-axis runs North-South, and the X-axis runs East-West, with North and East being positive. (a) True and estimated vehicle tracks from one Hydrone field experiment. The duration of the experiment is approximately 7 minutes. \blk{(b) True and estimated vehicle tracks from one Iver3 field experiment. The duration of the experiment is approximately 11 minutes.}}\label{fig:fieldTest}
    \vspace{-2mm}
\end{figure*}

Hydrone surface vehicle experiments took place in Mission Bay in San Diego, CA. There was a steady surface current to the south in addition to intermittent wave action on the surface from nearby boat activity. The vehicle was piloted by a remote operator, and the seafloor was flat and consisted of predominantly mud. These noise sources provide a great opportunity to study the robustness of our algorithm. Indeed, we found that the landmark sightings prevented significant estimation drift. Fig. \ref{fig:fieldTest}(a) shows results from one of three Hydrone experiments. To generate these results, we used the following inputs and system parameters. Inputs to the vehicle motion model are measured airspeed and heading; driving noise standard deviations $n_{\mathrm{s}}$, $n_{\mathrm{t}}$, $n_{\theta}$, and $n_{\gamma}$ estimated from the data were 0.1, 0.1, 1.5, and 0.25 respectively. Measurements for the update step were taken from the sonar sensor and the compass. Values used for noise terms $n_{d}$, $n_{\theta}$, and $n_{\gamma}$ were 0.75, 0.1 and 0.25 respectively. The compass measurements were adjusted with the declination appropriate for San Diego, CA. For \ac{mou} parameters we used $P_{\mathrm{det}}$ = 0.95, $\mu_{\mathrm{c}}$ = 0.01, and a clutter \ac{pdf} $f_{\mathrm{c}}(\V z_{k,l})$ that is a uniform \ac{pdf} between $-r_\mathrm{max}$ and $r_\mathrm{max}$. An onboard GPS sensor determined the true vehicle state and $r_\mathrm{max}$ was 20 meters.

\blk{Iver3 underwater vehicle experiments took place at the Scripps Institution of Oceanography in La Jolla, CA. The vehicle autonomously followed preset waypoints, remaining approximately 5 meters above the seafloor, which was flat and sandy. The data provided by the \ac{sss} on the Iver3 is significantly less noisy than the date provided by the \ac{sss} on the Hydrone. Since on the Iver3, the \ac{sss} is several meters below the surface, surface reflections were reduced significantly. The vehicle speed is consistent and maintained by the autonomous waypoint navigation system, reducing variability in driving speed and turn angle. There was very little subsurface current at the Scripps Pier compared to a strong surface current in Mission Bay. Differences in parameters settings and input of the proposed method compared to Hydrone experiments are as follows: Inputs to the vehicle motion model are measured speed-over-ground and heading; driving noise standard deviations $n_{\mathrm{s}}$, $n_{\mathrm{t}}$, $n_{\theta}$, and $n_{\gamma}$ estimated from the data were 0.2, 0.5, 1.5, and 1.5 respectively. The speed-over-ground is computed from DVL data. Measurements for the update step were taken from the sonar sensor and the compass. Values used for noise terms $n_{d}$, $n_{\theta}$, and $n_{\gamma}$ were 1.0, 0.1 and 0.25 respectively. The maximum sonar range $r_\mathrm{max}$ was set to 30 meters. Finally, a proprietary localization method from \textit{L3Harris} was used to determine the true vehicle state.}

\begin{figure*}[!htb]
    \centering
    \subfloat[]{\resizebox{0.48\linewidth}{!}{  \includegraphics[scale=.7]{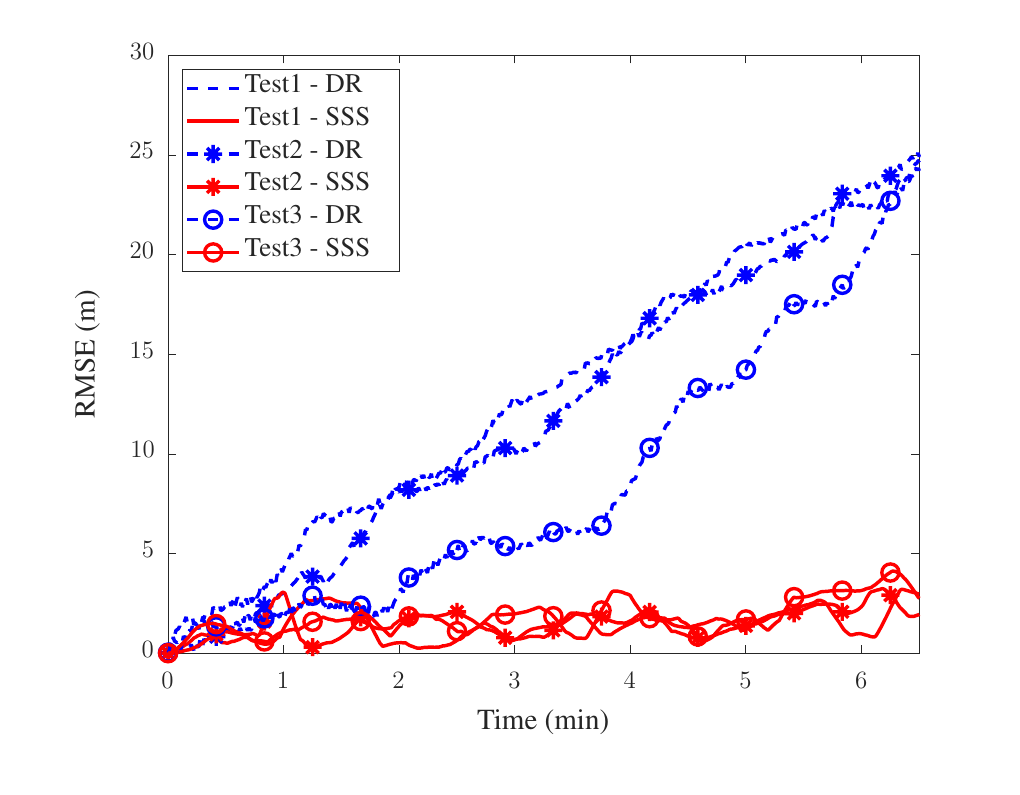} }} \hspace{1mm}
    \subfloat[]{\resizebox{0.49\linewidth}{!}{  \includegraphics[scale=.7]{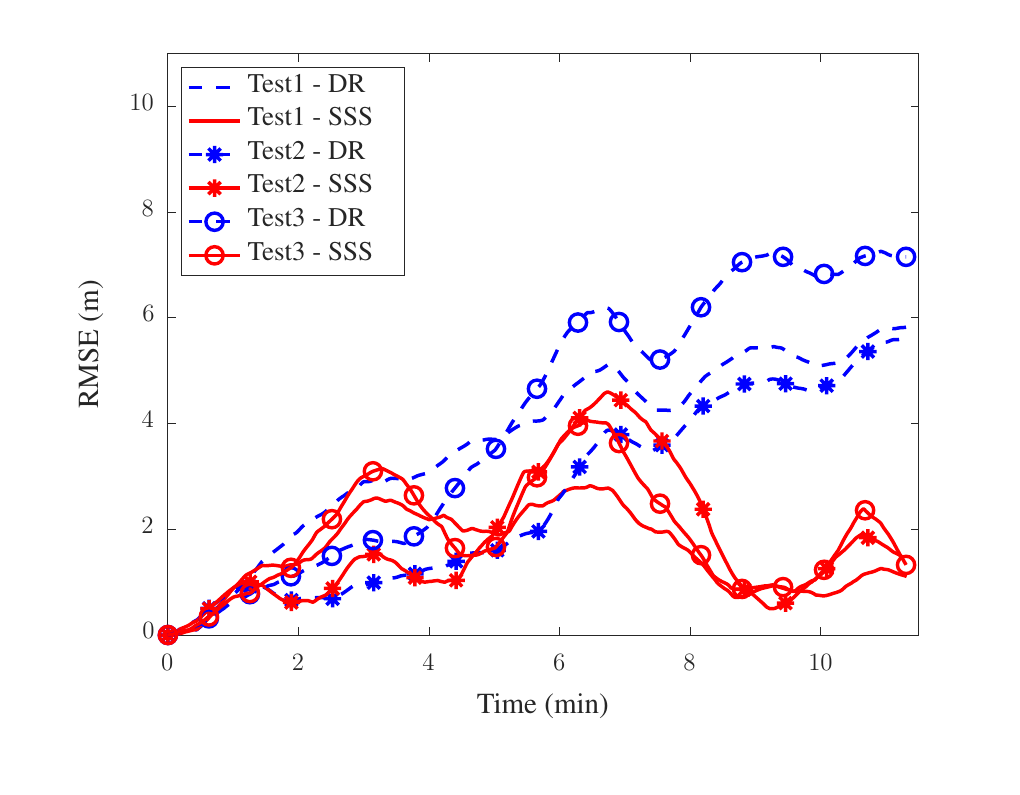} }}
    \caption{(a) Average \ac{rmse} versus time for position estimates from the navigation filter. Results from the three Hydrone field experiments are shown. The \ac{rmse} of the \ac{dr} method is shown for each experiment along with \ac{rmse} from the filter with \ac{sss} landmarks. \blk{(b) Same as (a) but for the three Iver3 field experiments.}}
    \label{fig:fieldError}
    \vspace{-2mm}
\end{figure*}

\blk{Fig. \ref{fig:fieldTest} shows two example field experiments, one with the Hydrone (a) and one with the Iver3 (b). The estimated and true vehicle location is shown in Cartesian coordinates relative to the vehicle starting point. In the case of the Hydrone surface vehicle, the \ac{dr} solution quickly drifts to the north, because there is a significant amount of surface current. Without absolute position information the \ac{dr} method does not adjust appropriately and the total error reaches 25 meters after 7 minutes. On the other hand, in the low noise Iver3 scenario, the \ac{dr} method performs much better. Dead reckoning still leads to steady drift over time, but the error grows more slowly. Compared to 25 meters in scenario 1, the \ac{dr} error in the second scenario is only about 7 meters after 11 minutes. In the Iver3 case, the \ac{sss} landmarks still bound the vehicle location error and outperform the \ac{dr} method, but the improvement is relatively smaller than with the Hydrone. The comparison of these results highlights the fact that absolute positioning with \ac{sss} landmarks is especially valuable when the environment is dynamic (e.g. fast or variable currents).}

\begin{figure*}[!htb]
    \centering
    \subfloat[]{\resizebox{0.48\linewidth}{!}{  \includegraphics[scale=.7]{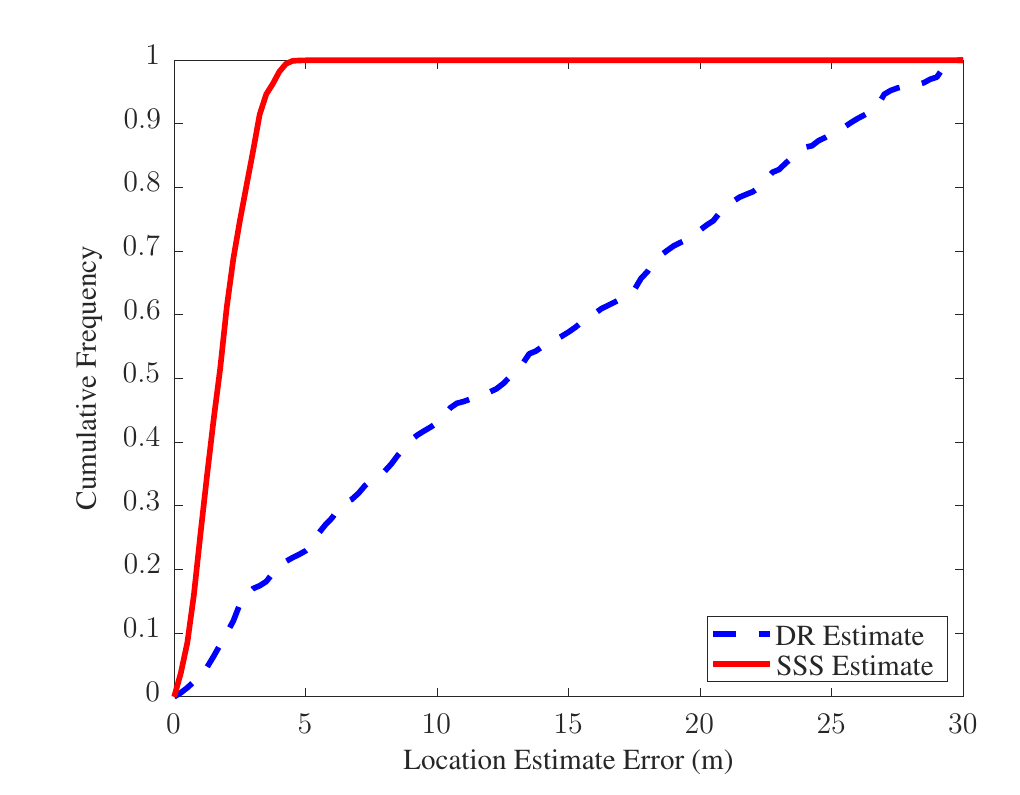} }} \hspace{1mm}
    \subfloat[]{\resizebox{0.49\linewidth}{!}{  \includegraphics[scale=.7]{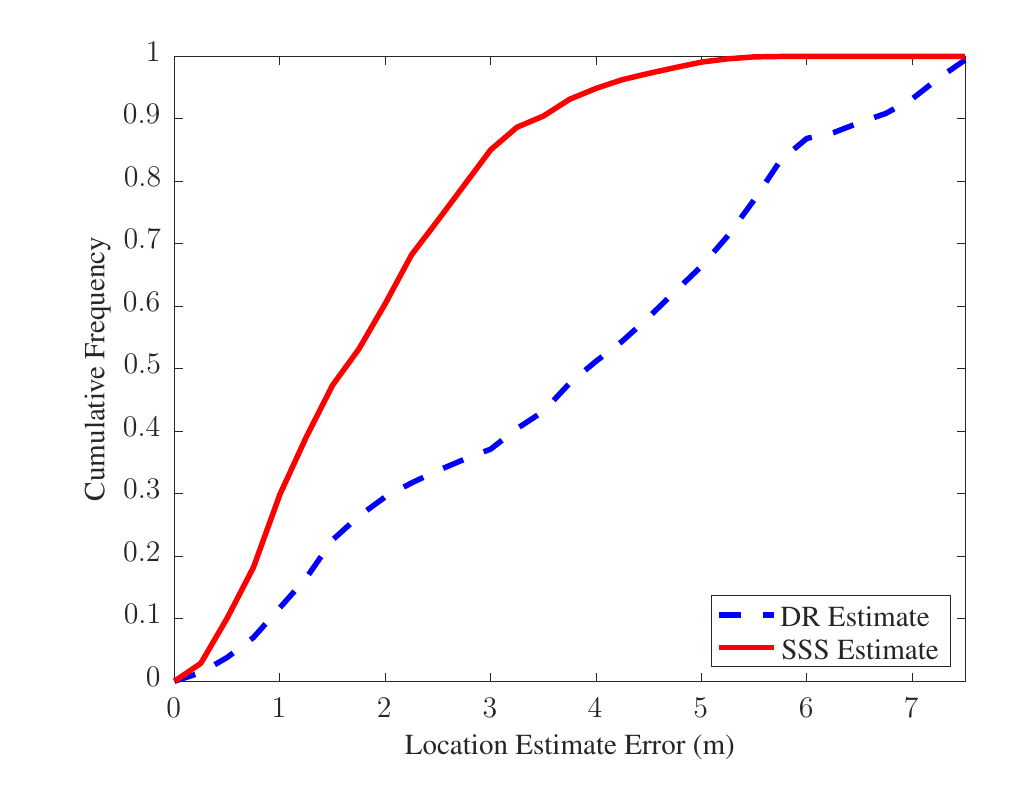} }}
    \caption{\blk{Cumulative Frequency (CF) plots for (a) Hydrone and (b) Iver3 experiments. These plots show the probability of the \ac{rmse} being below a certain error threshold at any given time step. The X-axis shows the \ac{rmse} threshold, and the Y-axis shows the probability that an estimate is below the error threshold. The SSS curves (red) indicate that at any given time, vehicle error remains below approximately 5m. In both cases the DR curves (blue) indicate that the number of estimates below a given error increases linearly with time. As can be seen in (b), it is clear that the DR error increases slower in the Iver3 case.}}
    \label{fig:CF_plot}
    \vspace{-2mm}
\end{figure*}

Fig. \ref{fig:fieldError} shows \ac{rmse} for all field experiments. In each case, the error with landmarks is clearly bounded and the error without increases indefinitely. The mean \ac{rmse} when using \ac{sss} landmarks was approximately 1.8 meters for the Hydrone scenario and \blk{1.7 meters for the Iver3 scenario. The error for the Iver3 \ac{sss} increases at a time of 3 minutes and again at a time of 6 minutes, which is attributed to the vehicle not seeing any landmarks for several minutes. This corresponds to the edges of the lawnmower pattern where the vehicle is far from the artificial landmarks. Once a landmark is spotted, the \ac{sss} error reduces dramatically and the \ac{dr} estimation error continues to grow. The probability of seeing a landmark was approximately $10\%$ in the Hydrone scenario and $2\%$ in the Iver3 scenario. These probabilities are similar to the values of $10\%$ and $1\%$ employed for the generation of synthetic data used for the results in Fig. \ref{fig:error}. Sparse landmarks can still limit vehicle location error in calm, low-noise environments as demonstrated by the Iver3 field experiments. In the La Jolla scenario, the \ac{dr} estimation method performs similarly as in simulation, whereas in the Mission Bay scenario, the \ac{dr} method performs significantly worse. We attribute this primarily to the wind and tide-driven surface currents in Mission Bay.} Finally, Fig. \ref{fig:CF_plot} shows \ac{cf} plots indicating the probability of the \ac{rmse} being below a certain error threshold at any given time step. The X-axis shows the \ac{rmse} threshold, and the Y-axis shows the probability that an estimate is below that error threshold.  The \ac{cf} plots are obtained by averaging over all time steps and all field experiments in each scenario. The presented results indicate that the \ac{rmse} is bounded around 5m when using \ac{sss} landmarks (for both scenarios) and that the \ac{dr} error is unbounded although increases at a different rate depending on environmental conditions.

\subsection{\blk{Computational Requirements}}
\label{sec:computation}

\begin{figure}[h]
\begin{center}
  \includegraphics[scale=.55]{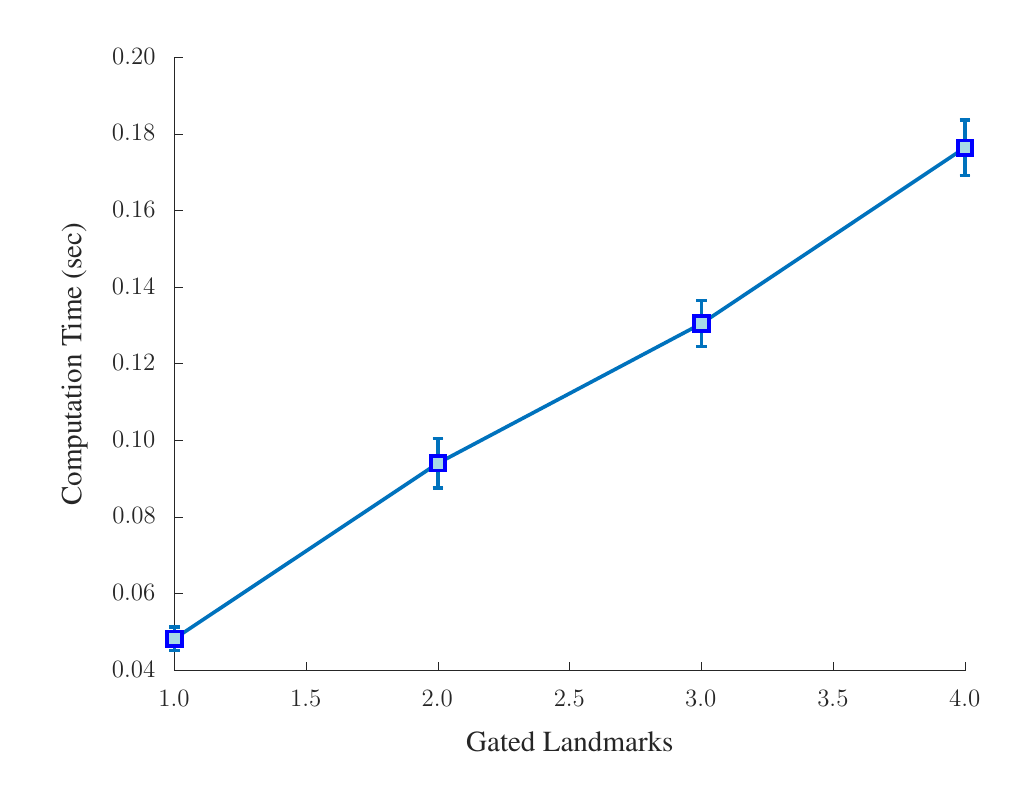}
\end{center}
  \caption{\blk{Average computation time per update step versus number of gated landmarks.}}
  \label{fig:landmark_computation}
\label{dnoise}
\end{figure}

\blk{In simulation, the average number of gated landmarks decreased rapidly with landmark spacing. In the case where landmarks were observed approximately $10\%$ of the time, the number of gated landmarks was 1.2 on average and had a maximum of 7. In contrast, in the case where the probability of seeing a landmark was $1\%$, the number of gated landmarks was 0.9 on average and had a maximum of 2. We processed the data from field experiments to determine the average time to complete an update step as a function of the number of gated landmarks. As can be seen in Fig. \ref{fig:landmark_computation}, this relationship is roughly linear. As discussed in Sec. \ref{sec:complexity}, in a worst case scenario, computation time scales quadratically with the number of gated landmarks. In the considered scenario, computation time scales almost linearly because the number of measurements does not increase significantly with the number of gated landmarks. In general, computation time will depend on the maximum range of the sonar and the size of the gated region.}

\blk{All simulation and field experiments use 10,000 particles in the update step. On an Apple M3 MAX with 36GB of RAM and a 14-core CPU, the average time to complete a simulation update step is 0.004 seconds and 0.003 seconds when the probability of seeing a landmark was $10\%$ and $1\%$, respectively. The average time per update step is 0.004 seconds in the Hydrone case, and 0.002 seconds in the Iver3 case. These values closely match the computation times from the simulation. Running the proposed navigation method on a 12-minute Iver3 experiment with a 30Hz sonar (13,000 time steps) takes approximately 30 seconds. These results demonstrate that our method is significantly faster than the sonar update rate, and it can thus be performed in real-time. The update step is at least one order of magnitude computationally more expensive than the prediction step, so we do not include an analysis of the prediction step here. An important point, however, is that this update step does not include automatic detection of the landmarks in the \ac{sss} images, which will still need to be added in a real-time system. The landmark-aided navigation method proposed here is efficient and leaves computational room for automatic detection to be incorporated.}

\section{Conclusion and Future Work}
\label{sec:Conclusion}

We developed and validated a method that increases navigation accuracy for \acp{sauv} with \ac{sss} and has potential for real-time application.  The error of the vehicle location estimate is bounded by incorporating landmark detections from \ac{sss} data and using individual pings as measurements. Our simulation results indicated an improved navigation performance compared to a reference method that relies on \ac{dr} in various scenarios. These simulation results showed unreasonably low \ac{rmse} compared to real-world scenarios but allowed us to assess the effect of landmark spacing. \blk{We demonstrated significant improvements in state estimation accuracy when applying our method to data collected in field experiments in a dynamic environment. In addition, we demonstrated moderate improvement in a quiescent environment.} The method is computationally efficient and statistically robust, making it possible to extend its use to real-time applications at sea. It is worth noting that the towing \ac{sss} near the seafloor, rather than on the surface, generates cleaner \ac{sss} returns to use for landmark detection.

The following steps of this research consist of (i) automatic landmark detection via embedded deep neural networks \cite{LiaMey:J23}, (ii) estimation of vehicle height from the sea floor using the image nadir, and (iii) reseeding the vehicle proposal \ac{pdf} after long periods without landmark sightings. Task (iii) will depend on task (i) as it requires labeling or classification of landmarks and cannot be performed solely with detection. There are several documented methods to achieve task (i), as discussed in Section I.  One approach that may be generalized to different seabeds is the application of unsupervised learning in the form of anomaly detection \cite{9705947}. A self-supervised approach used for landmark detection from aerial images could also be applied to \ac{sss} images, where contrastive learning is used with a convolutional neural network to discern whether two cropped images were captured from the exact location \cite{lee4self}. These latter approaches can be more computationally taxing but have the potential advantage of robustness across environments. Task (ii) is currently being pursued using advanced image processing methods and ideas from \cite{RawElmFraCurYua:C17}. The completion of task (iii) will be important in scenarios where a landmark is detected but the posterior \ac{pdf} of the vehicle state has drifted and is too uninformative to accurately associate the detection. Instead of using the predicted posterior as the proposal \ac{pdf} in the update step, an alternative proposal \ac{pdf} will be constructed based on the details of the detection and the known \ac{pdf} of landmarks. \blk{The aforementioned steps describe a complete navigation system for \textit{known} environments. To obtain a navigation system that works in \textit{unknown} environments, the map of landmarks will need to be augmented in real-time in a \ac{slam} approach. In addition, the probabilistic framework could be expanded to include dynamic landmarks that may change over time.}

\selectfont
\bibliographystyle{IEEEtran}
\bibliography{IEEEabrv,StringDefinitions,Books20,Papers20,Temp}

\end{document}